\pdfoutput=1

\documentclass[11pt]{article}

\usepackage[final]{acl}

\usepackage[most]{tcolorbox}
\usepackage{multicol}
\tcbuselibrary{raster,skins,breakable,listings}
\usepackage{times}
\usepackage{latexsym}
\usepackage{hyperref}
\usepackage{makecell}
\usepackage{stfloats}
\usepackage{ulem} 

\usepackage[most]{tcolorbox}
\tcbuselibrary{skins, breakable, raster}
\tcbset{enhanced, breakable}
\usepackage{color}
\usetikzlibrary{patterns}      
\usepackage{tikz-qtree}        
\usetikzlibrary{external}
\usepackage[T1]{fontenc}

\usepackage[utf8]{inputenc}

\usepackage{url}
\usepackage{listings}
\usepackage[ruled,vlined]{algorithm2e}
\usepackage{algpseudocode}
\usepackage{inconsolata}
\usepackage{xcolor}

\usepackage{amsmath}
\usepackage{amssymb}
\usepackage{graphicx}
\usepackage{subcaption} 
\usepackage{booktabs}
\usepackage{adjustbox}
\usepackage{multirow}
\usepackage{ulem}

\definecolor{fontgray}{RGB}{44, 62, 80}
\definecolor{myred}{RGB}{235, 47, 6} 
\definecolor{summertime}{RGB}{245, 205, 121}
\definecolor{darkgrass}{RGB}{0, 148, 50}
\definecolor{myblue}{RGB}{0, 168, 255}
\definecolor{mygray}{RGB}{158, 158, 158}
\definecolor{puffin}{RGB}{250, 152, 58}
\definecolor{lowpurple}{RGB}{210, 180, 222}
\definecolor{lowblue}{RGB}{102,178,255}
\definecolor{lowred}{RGB}{245, 183, 177}
\definecolor{deeppurple}{RGB}{142, 68, 173}
\definecolor{nephritis}{RGB}{39, 174, 96}

\definecolor{deepblue}{RGB}{41, 128, 185}
\definecolor{shymoment}{RGB}{162, 155, 254}
\definecolor{firstdate}{RGB}{250, 177, 160}
\definecolor{mintleaf}{RGB}{0, 184, 148}
\definecolor{alizarin}{RGB}{231, 76, 60}
\definecolor{soaring}{RGB}{149, 175, 192}
\definecolor{electronblue}{RGB}{9, 132, 227}
\definecolor{pinkgla}{RGB}{0, 184, 148}
\definecolor{coral}{RGB}{255, 127, 80}

\lstdefinestyle{python}{
    language=Python,
    basicstyle=\fontsize{7}{8}\ttfamily,
    keywordstyle=\color{blue},
    commentstyle=\color{gray},
    stringstyle=\color{black},
    showstringspaces=false,
    breaklines=true,
    breakindent=0pt,
    breakatwhitespace=false,
    escapeinside={(*@}{@*)},
    moredelim=[is][\bfseries]{(*@}{@*)}
}

\lstdefinestyle{plain}{
    basicstyle=\fontsize{8}{10}\ttfamily,
    commentstyle=\color{gray},
    stringstyle=\color{green},
    showstringspaces=false,
    breaklines=true,
    breakatwhitespace=false,
    breakindent=0pt,
    escapeinside={(*@}{@*)},
    literate={_}{{\_}}1
}

\title{Parrot: A Training Pipeline Enhances Both Program CoT and Natural Language CoT for Reasoning}

\author{
 \textbf{Senjie Jin\textsuperscript{1}\thanks{{ }\ Equal contribution.$^\dagger$Corresponding authors.}},
 \textbf{Lu Chen\textsuperscript{1}$^{*}$},
 \textbf{Zhiheng Xi\textsuperscript{1}$^{*}$},
 \textbf{Yuhui Wang\textsuperscript{1}},
\\
 \textbf{Sirui Song\textsuperscript{1}},
 \textbf{Yuhao Zhou\textsuperscript{1}},
 \textbf{Xinbo Zhang\textsuperscript{2}$^\dagger$},
 \textbf{Peng Sun\textsuperscript{2}},
\\
 \textbf{Hong Lu\textsuperscript{1,4}},
 \textbf{Tao Gui\textsuperscript{1,3,4}$^\dagger$},
 \textbf{Qi Zhang\textsuperscript{1,4}},
 \textbf{Xuanjing Huang \textsuperscript{1,4}}
\\
 \textsuperscript{1}College of Computer Science and Artificial Intelligence,
Fudan University \\
 \textsuperscript{2}ByteDance Research 
 \textsuperscript{3}Shanghai Innovation Institute \\
 \textsuperscript{4}Shanghai Key Laboratory of Intelligent Information Processing
\\
 \texttt{sjjin24@m.fudan.edu.cn} \\
 \texttt{zhangxinbo.freya@bytedance.com, tgui@fudan.edu.cn}\\
}

\usepackage{tcolorbox}  
\tcbuselibrary{skins}   
\tcbuselibrary{breakable} 

\begin{document}
\maketitle
\begin{abstract}
Natural language chain-of-thought (N-CoT) and Program chain-of-thought (P-CoT) have emerged as two primary paradigms for large language models (LLMs) to solve mathematical reasoning problems. Current research typically endeavors to achieve unidirectional enhancement: P-CoT enhanced N-CoT or N-CoT enhanced P-CoT. In this paper, we seek to fully unleash the two paradigms' strengths for mutual enhancement and ultimately achieve simultaneous improvements.
We conduct a detailed analysis of the error types across two paradigms, based on which we propose \textbf{Parrot}, a novel training pipeline for mathematical problems: 1) Three target-designed subtasks integrate sequential P-CoT and N-CoT generation. 2) A subtask hybrid training strategy to facilitate natural language semantic transferability. 3) The converted N-CoT auxiliary reward is designed to alleviate the sparse rewards in P-CoT optimization.
Extensive experiments demonstrate that Parrot significantly enhances both the performance of N-CoT and P-CoT, especially on N-CoT. Using Parrot SFT, the LLaMA2’s and CodeLLaMA’s N-CoT performance achieve gains of +21.87 and +21.48 on MathQA over the RL baseline, which is resource-intensive\footnote{https://github.com/Leonnnnnn929/ParrotTraining}.


\end{abstract}



\section{Introduction}
Large language models (LLMs) have exhibited an impressive success in multi-step mathematical reasoning \cite{DBLP:conf/acl/WangLSXDLCWS24, DBLP:journals/corr/abs-2402-03300, DBLP:conf/icml/WanFWM00024}. 
The existing work primarily concentrates on enabling models to generate natural language chain-of-thought (N-CoT) rationales \cite{wei2022chain} or leverage executable and verifiable code, such as Python \cite{chen2022program, gao2023pal, luong2024reft, xi2024training}, to generate program chain-of-thought (P-CoT) 
for offloading intensive calculations \cite{li2024evaluating}. These two paradigms exhibit distinct advantages. Specifically, N-CoT introduces more reasoning details by an explicit thinking process \cite{lin2024lean}, which is more comprehensible and holds a broader applicability \cite{DBLP:journals/corr/abs-2405-06682, DBLP:journals/corr/abs-2409-12917}, while P-CoT demonstrates high effectiveness \cite{gao2023pal} and enables easy process verification \cite{gou2023tora}. 

Current research typically endeavors to utilize one to facilitate the other: (1) N-CoT-enhanced P-CoT. Integrating an explicit natural language analysis prior to each code step or the entire code solution \cite{gao2023pal, lin2024lean, li2024humans}. (2) P-CoT-enhanced N-CoT. Presenting specific procedures as code and invoking them through an external verifier \cite{gou2023tora}. Although \cite{DBLP:conf/iclr/YueQZFH00C24} proposes a N-CoT\&P-CoT rationale hybrid training strategy, which mainly aims at the solution diversity. The synergistic facilitation potential between these paradigms has not been sufficiently explored.

In this paper, we first conduct a comprehensive error analysis (Section \ref{Pre}) of these two paradigms and find that, on the one hand, in addition to intrinsic limitations in logical reasoning, the approach of directly generating P-CoT from problems struggles with accurate variable definition and problem comprehension \cite{DBLP:conf/iclr/YueQZFH00C24, li2024humans}. We integrate these capabilities suitable for natural language by constructing specialized subtasks and employing hybrid training. On the other hand, N-CoT mainly suffers from logical confusion \cite{DBLP:conf/emnlp/XiJZZGLGZH23, wang2022self} as well as calculation errors in intermediate steps \cite{gao2023pal}. We enable N-CoT to refer to the concise P-CoT reasoning steps and incorporate the intermediate results of the latter as a simple yet effective form of process supervision \cite{lightman2023let}.

\begin{figure*}[t]
    \includegraphics[width=0.95\linewidth]{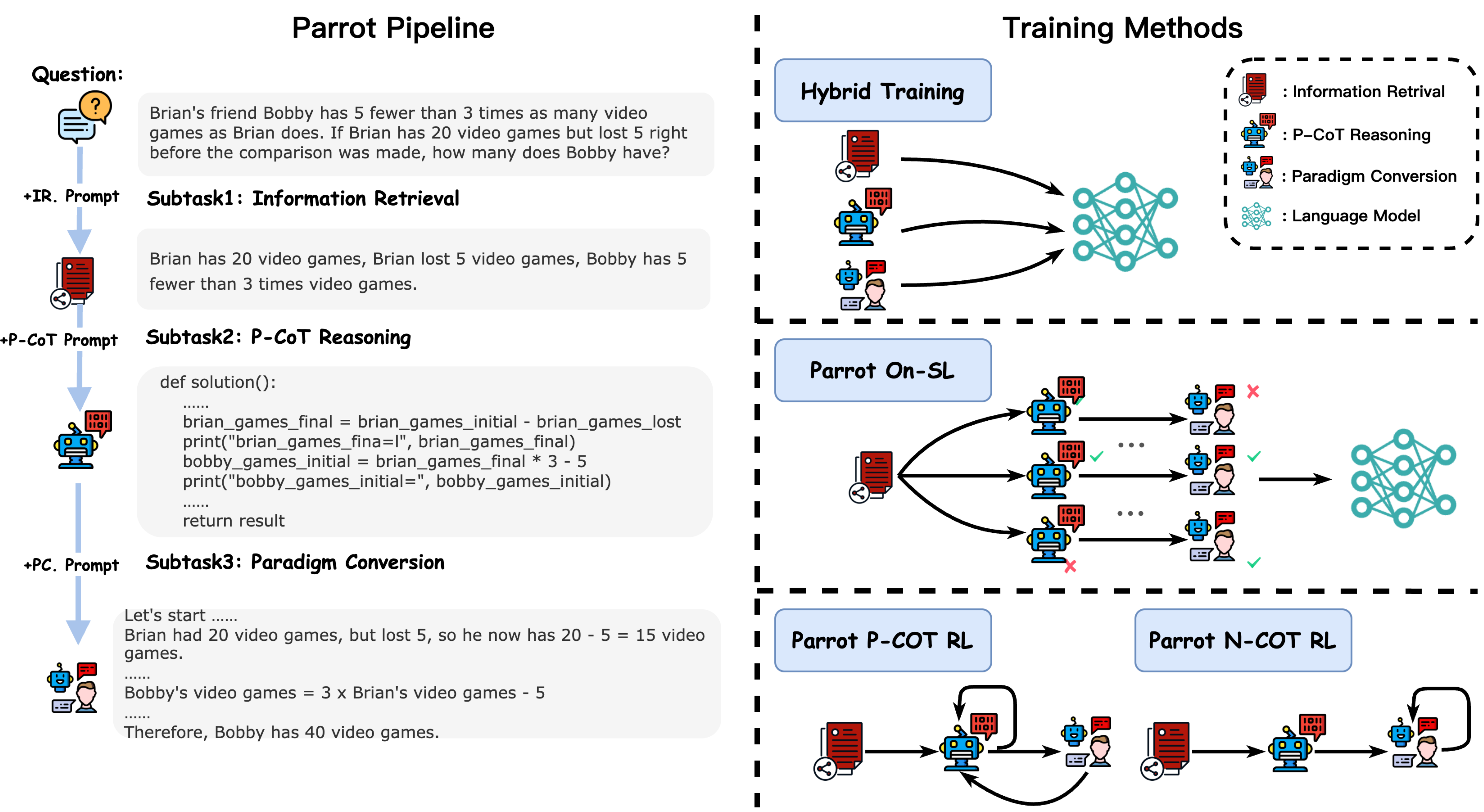}
    \centering
 	\caption{
 The training pipeline and methods of \textbf{Parrot}. On the left, the pipeline consists of three subtasks: \textbf{Information Retrieval}, \textbf{P-CoT Reasoning}, and \textbf{Paradigm Conversion}. By these subtasks, the model sequentially generates P-CoT and n-CoT. On the right, we use a Hybrid Supervised Fine-Tuning (SFT) strategy to enable semantic transfer and incorporate reinforced algorithms for further improvements. The detailed Parrot inference process and subtask prompts are provided in Appendix \ref{sub prom}.
  }\label{fig:main}
\end{figure*}

Based on the above, we propose \textbf{Parrot}, as illustrated in Figure \ref{fig:main}, a novel training pipeline to promote both P-CoT and N-CoT performance on mathematical problems. The pipeline comprises three target-designed subtasks: \textbf{Information Retrieval} trains the model to concentrate on key information within problem. \textbf{P-CoT Reasoning} utilizes the information to generate variable well-defined code solutions. \textbf{Paradigm Conversion} enhances N-CoT with concise P-CoT and its intermediate outputs. This pipeline also aligns with the human problem-solving process \citep{krawec2014problem}, which involves three stages: individuals examine the problem and identify key information, then utilize the formalized language for unambiguous declarations, thereby incorporating the characteristic problem context to generate interpretable and accessible resolutions. \cite{kazemi2012investigation}. 

Regarding methodology, we initially adopt a hybrid Supervised Fine-Tuning (SFT) strategy, enabling the model to master subtasks while enhancing P-CoT through transferability across remaining subtasks \cite{DBLP:conf/iclr/YueQZFH00C24}.
We will thoroughly discuss the impact of each sub-task in the analysis section \ref{sec: sub ab}. Furthermore,
we introduce Reinforcement Learning (RL) to verify Parrot’s applicability under different fine-tuning methods and data efficiency.
During Online Self-Learning (On-SL) \cite{DBLP:journals/corr/abs-2211-14275, DBLP:conf/nips/AnthonyTB17}, we collect N-CoT solutions and use them in SFT to demonstrate their quality with the support of P-CoT.
In the Proximal Policy Optimization (PPO) \cite{schulman2017proximal} stage, we use the validity of the converted N-CoT as the auxiliary reward signal to mitigate the issue of sparse rewards \cite{DBLP:journals/corr/abs-1709-00103, le2022coderl} for P-CoT verification in mathematical reasoning.


In summary, we make the following contributions:

(1) We carry out a comprehensive analysis of limitations for coding-expertise (CodeLLaMA) and non-coding-expertise (LLaMA2) within P-CoT and N-CoT paradigms.

(2) We propose Parrot, a novel training pipeline enhancing both P-CoT and N-CoT mathematical reasoning performance. Additionally, we conduct extensive ablations to analyze the impact of each sub-task.

(3) We perform SFT on the collected N-CoT from On-SL to validate its quality with the aid of P-CoT, and we use the N-CoT auxiliary reward to mitigate the reward sparsity issue in the P-CoT RL phase.


(4) We conduct extensive experiments on three difficulty-level datasets and model families, which indicate that Parrot can effectively improve the model’s P-CoT and N-CoT reasoning performance, especially on N-CoT.


\begin{figure}[t]
    \centering
    \hspace*{-0.7cm}  
    \includegraphics[width=1.2\linewidth]{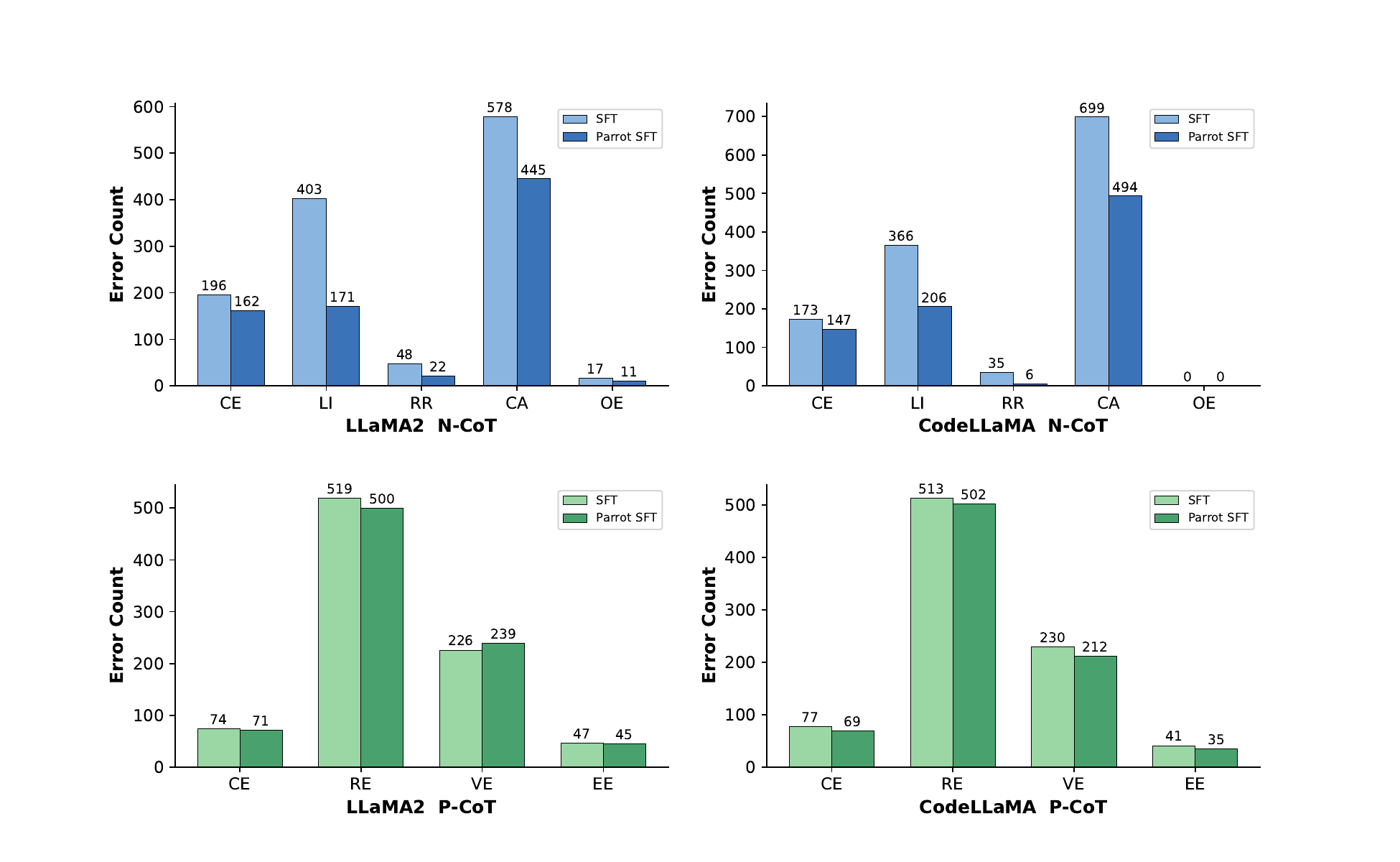}
    \caption{The histogram of error types. The labels on the x-axis are defined in section \ref{Define}, while \textbf{OE} denotes Other Errors. Results from SFT are shaded in light colors, and Parrot SFT results are presented in dark colors.}
    \label{fig:error type}
\end{figure}

\section{Preliminary Analysis}\label{Pre}
Pre-training on different corpus compositions \cite{lu2024mathcoder2} and reasoning paradigms collectively determines the model performance across math problems. We first perform a detailed error analysis on the coding-expertise models (CodeLLaMA) \cite{DBLP:journals/corr/abs-2308-12950} and non-coding-expertise (LLaMA2) models \cite{DBLP:journals/corr/abs-2307-09288} to investigate their intrinsic limitations in P-CoT and N-CoT. Following previous work \cite{luong2024reft}, we perform Supervised Fine-Tuning (SFT) training on the MathQA \cite{DBLP:conf/naacl/AminiGLKCH19} dataset and collect error samples. The error types and analysis are elaborated in the following sections.


\subsection{Empirical Identification of Error Types} \label{Define}
We first randomly sampled 50 error cases from each paradigm for manual examination. Our findings reveal that: (1) For N-CoT, except \textbf{calculation error} \cite{gao2023pal}, the model also suffers from \textbf{logical inconsistency}, \textbf{problem comprehension}, \textbf{redundant and repetitive} information \cite{li2024humans}. (2) For P-CoT, besides the model's inherent \textbf{reasoning} limitations, generating P-CoT directly from problems has an issue in \textbf{problem comprehension}, \textbf{variable definition}, and \textbf{expression error}. We then utilize GPT-4 \cite{DBLP:journals/corr/abs-2303-08774} to statistically analyze all error samples. Specific descriptions of error types and evaluation prompts are detailed in the Appendix \ref{appendix:prompts}, and error examples in Appendix \ref{example error}.


\subsection{Error Analysis}
The statistical results are shown in Figure \ref{fig:error type}. We observe that: (1) For N-CoT, consistent with previous work \cite{gao2023pal}, approximately half of the errors stem from calculation (\textbf{CA}), followed by logic inconsistency (\textbf{LI}), which we hypothesize are due to the absence of process supervision. In some cases, the model also exhibits the phenomenon of redundant and repetitive (\textbf{RR}). From the model perspective, incremental training in code enhances the model's logical capabilities and reasoning conciseness. However, this also results in insufficient semantic understanding, leading to more calculation errors. (2) For P-CoT, as proposed by \cite{li2024humans}, the code is inferior to natural language in semantic analysis and abstract reasoning. The primary issue arises from reasoning errors (\textbf{RE}). The model also fails short on variable definition errors (\textbf{VE}), accounting for one-quarter errors and Expression Errors (\textbf{EE}). Both paradigms struggle with problem comprehension errors (\textbf{CE}), with P-CoT fewer since its variable definition analysis.




\section{Method}
\paragraph{Motivation.}
From current cross-party facilitating works and the error types uncovered in section \ref{Pre}, 
we aim to explore the feasibility of leveraging strengths to mitigate counterpart drawbacks in P-CoT and N-CoT and ultimately achieve collective performance improvement.
Hence, we propose \textbf{Parrot}, a novel training pipeline that focuses on key information to facilitate the variable well-defined P-CoT and generates N-CoT based on P-CoT and its intermediate outputs. We elaborate the details about pipeline subtasks in section \ref{subtask} and training methods in section \ref{sec: hb training} and section \ref{RL}, which is illustrated in Figure \ref{fig:main}.
\subsection{Pipeline Subtask Construction}\label{subtask}
We organically decompose the full training pipeline into three targeted distinct subtasks. 1) \textbf{Information Retrieval}. Although P-CoT enables precise calculations with the support of the external verifier, it often suffers from erroneous variable definitions \cite{jie2023design}. We first orient the model’s attention on key numeric information within the problem to achieve variable well-defined in P-CoT. For a given problem $\textit{x}$ and information retrieval prompt $p_{1}$, key information $d_{1}$ is generated by:

\begin{equation}\label{eq: numerics retrieval}
\small
d_{1} \sim \Pi(\cdot|x \oplus p_{1}),
\end{equation}

where $\oplus$ donates concatenation. \textbf{Note} in this phase, no extra knowledge is incorporated. 
2) \textbf{P-CoT Reasoning}. Subsequently, utilizing the key information $d_{1}$ and code inference prompt $p_{2}$, the model generates a python snippet $d_{2}$:

\begin{equation}\label{eq: formal reasoning}
\small
d_{2} \sim \Pi(\cdot|x \oplus p_{1} \oplus d_{1} \oplus p_{2}),
\end{equation}

which is then validated by invoking the interpreter. 3) \textbf{Paradigms Conversion}. By harnessing the model's multilingual alignment capability \cite{DBLP:conf/iclr/Xu0SA24}, we generate more understandable and widely accessible N-CoT based on the P-CoT, its intermediate results $i$ and prompt $p_{3}$:

\begin{equation}\label{eq: resolution conversion}
\small
d_{3} \sim \Pi(\cdot|x \oplus p_{1} \oplus d_{1} \oplus p_{2} \oplus d_{2} \oplus i \oplus p_{3}),
\end{equation}

as analyzed in section \ref{Pre}, there are mainly two reasons for this N-CoT generation strategy: Firstly, the code reasoning features concise steps, which help alleviate repetition and redundancy errors. Secondly, the main errors with N-CoT are calculations and logical inconsistencies. Beyond precise calculations, incorporating P-CoT’s intermediate results can serve as a simple and effective process supervision \cite{luo2024improve, chen2025better}, and ablation analysis in section \ref{N-CoT} validates our hypothesis.

\subsection{Subtask Hybrid Training} \label{sec: hb training}
Motivated by \cite{DBLP:conf/iclr/YueQZFH00C24},
we adopt a hybrid training strategy and structure all the subtasks into a unified input-output form to perform multi-task SFT training \cite{zhang2021survey} instead of training sequentially by subtasks, which often involves the challenge of knowledge degradation \cite{DBLP:conf/emnlp/XuQGW0ZX24, DBLP:conf/nips/SuZQ0LSLZC24} and impairs the model's performance. 
This strategy is poised to facilitate problem comprehension due to the incorporation of solution diversity from P-CoT and N-CoT \cite{liang2024improving}, and to transfer the explicit reasoning traces from N-CoT for semantic analysis \cite{lin2024lean}, ultimately to enhance the logical reasoning ability of P-CoT.
We conduct extensive ablation experiments to validate our hypotheses and thoroughly analyze the impact of each subtask within the pipeline in section \ref{sec: sub ab}.
 
\subsection{Reinforcement Enhanced Reasoning}\label{RL}
Upon completing model initialization through hybrid training, we incorporate reinforcement learning algorithms to further verify Parrot’s applicability under different fine-tuning methods and data efficiency.
\paragraph{Online Self-learning.}
We implement the online self-training (On-SL) following \cite{luong2024reft}. In our setup, the model sequentially generates P-CoT and N-CoT rollouts, using jointly correct samples to augment training with the original datasets.
\paragraph{Proximal Policy Optimization.}
We leverage proximal policy optimization (PPO) \cite{schulman2017proximal} with a clipped objective as reinforcement learning (RL) algorithm. The final token before <eos> of the sampled sequence is assigned a reward score, while all remaining tokens receive 0 \cite{yu2023outcome, xi2024training}.

\begin{equation}\label{eq: reward function1}
\small
R(s_{t-1},a_{t}) = \left\{  \begin{aligned} 
& R_{f}(s_{t-1},a_{t}),  \ \ &t = T \\
& 0, \ \ &t \neq T
\end{aligned} 
\right.,
\end{equation}

where $R_f(\cdot)$ is a rule-based reward function merely relies on the correctness of the answer. Despite its efficiency, it suffers reward sparsity. Inspired by partial reward design \cite{li2024humans, le2022coderl}, 
we use the validity of the converted N-CoT as the auxiliary reward signal to verify the P-CoT:

\begin{equation}\label{eq: reward function2}
\small
R_f(s_{T-1},a_{T}) = \left\{  \begin{aligned} 
1,  \ \ \ \  & \text{Both answer correct}  \\
1 - \gamma,  \ \ \ \  & \text{P-CoT correct, N-CoT null} \\
\epsilon, \ \ \ \ & \text{P-CoT not, but numeric}  \\
0,\ \ \ \    & \text{P-CoT null}
\end{aligned}
\right. 
\end{equation}

when the converted N-CoT is incorrect but of numeric type, we consider it a calculation error. For cases with no answer, we give P-CoT reasoning a penalty $\gamma$ for comprehension difficulty to enhance its effectiveness. The value model $V_{\mathit{\Phi}}$ is constructed by appending a linear value head on top of the last hidden states of the policy model $\pi_{\theta}^{p}$. Consistent with \cite{luong2024reft}, the final reward $R_{f}(s_{t-1},a_t)$ integrates both the reward score and the token-level Kullback-Leibler (KL) divergence \cite{kullback1951information}.
Based on the reward $R$ and value model $V_{\mathit{\Phi}}$, we estimated the generalized advantage esti-mate (GAE) \cite{schulman2017proximal} $A(s_{t-1},a_t)$, and the optimal objective is to maximize the return:

\begin{equation}
\small
\begin{aligned} \label{eqn: policy gradient}
     \mathbb{E}_{\tau \sim \pi^{p}_\theta} \left[ \sum_{t=1}^T   \nabla_\theta \log \pi^{p}_\theta(a_{t}|s_{t-1}) A(s_{t-1},a_t) \right],
\end{aligned}
\end{equation}

where $\tau$ is the sampled sequence. 


\begin{table*}[t]
\small
    \centering
    \adjustbox{max width=1.0\linewidth}{
    \begin{tabular}{l|ccccccc|cc}
        \toprule
        \multirow{2}{*}{\bf Training Method} & \multirow{2}{*}{\bf Size} & \multicolumn{2}{c}{\bf GSM8K} & \multicolumn{2}{c}{\bf SVAMP} & \multicolumn{2}{c}{\textbf{MathQA}$_\text{numeric}$} & \multicolumn{2}{c}{\bf Average}\\
        &  & \textbf{N-CoT} & \textbf{P-CoT} & \textbf{N-CoT} & \textbf{P-CoT} & \textbf{N-CoT} & \textbf{P-CoT} & \textbf{N-CoT} & \textbf{P-CoT}  \\
        \midrule
Tora + CodeLLaMA & 7B                                      & - & $72.60^*$  & - & $70.40^*$ & - & - & - & -  \\
MathGenie + LlaMA2 & 7B                                      & - & $71.70^*$  & - & $78.50^*$ & - & - & - & -  \\
MathGenie + CodeLLaMA & 7B                                      & - & $71.50^*$  & - & $80.20^*$ & - & - & - & -  \\
DotaMath + LlaMA2 & 7B                                      & - & $79.60^*$  & - & - & - & - & - & -  \\
MARIO + DeepSeek & 7B                                      & - & $78.40^*$  & - & - & - & - & - & -  \\
HTL + CodeLLaMA & 7B                                      & - & $65.70^*$ & - & $74.40^*$ & - & - & - & -  \\
HTL + Mistral & 7B                                      & - & $78.10^*$ & - & $82.40^*$ & - & - & - & -  \\
GPT-4 & -                                       & $92.72$ & $97.00^*$ & $91.60$ & $94.80^*$ & $83.17$ & $66.29$ & $89.16$ & $86.03$\\
        \midrule
        \midrule
LLaMA2 + SFT & 7B                                      & $44.05$ & $58.61$ & $58.60$ & $69.50$ & $22.62$ & $46.04$  & $41.76$ & $58.05$ \\
LLaMA2 + MAmmoTH SFT & 7B                              & $47.54$ & $58.15$ & $59.30$ & $\uline{71.90}$ & $27.28$ & $44.80$  & $44.71$ & $58.28$ \\
LLaMA2 + On-SL & 7B                    & $45.94$ & $60.80$ & $\textbf{60.70}$ & $69.40$ & $30.15$ & $46.48$  & $45.60$ & $58.89$ \\
LLaMA2 + RL  & 7B                     & $44.96$ & $\uline{63.99}$ & $59.70$ & $71.40$ & $26.92$ & $44.92$  & $43.86$ & $\uline{60.10}$ \\
LLaMA2 + Parrot SFT &  7B            & $60.81$ & $59.74$ & $59.60$ & $71.60$ & $48.79$ & $\uline{46.73}$  & $56.40$ & $59.42$ \\
LLaMA2 + Parrot On-SL &  7B            & $\uline{60.96}$ & $59.21$ & $59.40$ & $69.60$ & $\uline{49.22}$ & $45.92$  & $\uline{56.53}$ & $58.24$ \\
LLaMA2 + Parrot RL &  7B            & $\textbf{61.26}$ & $\textbf{66.03}$ & $\uline{60.00}$ & $\textbf{73.60}$ & $\textbf{50.37}$ & $\textbf{47.66}$  & $\textbf{57.21}$ & $\textbf{62.43}$ \\
        \midrule
        \midrule
CodeLLaMA + SFT & 7B                                      & $44.88$ & $65.05$ & $56.70$ & $75.50$ & $22.37$ & $47.04$  & $41.32$ & $62.53$ \\
CodeLLaMA + MAmmoTH SFT & 7B                              & $46.70$ & $65.50$ & $62.50$ & $75.70$ & $24.05$ & $46.23$  & $44.42$ & $62.48$ \\
CodeLLaMA + On-SL & 7B                    & $45.19$ & $65.43$ & $59.70$ & $76.30$ & $26.17$ & $48.10$  & $43.69$ & $63.28$ \\
CodeLLaMA + RL  & 7B                     & $53.22$ & $\uline{72.78}$ & $62.30$ & $\uline{78.40}$ & $25.36$ & $48.16$  & $46.96$ & $\uline{66.45}$ \\
CodeLLaMA + Parrot SFT &  7B            & $\uline{64.90}$ & $66.19$ & $\uline{62.90}$ & $77.60$ & $46.84$ & $\textbf{49.03}$  & $\uline{58.21}$ & $64.27$ \\
CodeLLaMA + Parrot On-SL &  7B            & $64.82$ & $65.73$ & $61.70$ & $75.40$ & $\uline{47.04}$ & $48.29$  & $57.85$ & $63.14$ \\
CodeLLaMA + Parrot RL &  7B            & $\textbf{65.04}$ & $\textbf{74.53}$ & $\textbf{64.30}$ & $\textbf{79.60}$ & $\textbf{48.35}$ & $\uline{48.85}$  & $\textbf{59.23}$ & $\textbf{67.66}$ \\
        \bottomrule
    \end{tabular}
    }
    \caption{The main experimental results on three benchmarks and two models. We simultaneously evaluated the model's performance on N-CoT and P-CoT. The results of Parrot-based methods are presented at the bottom of each block, with the overall performance outperforming those corresponding baselines. The best result is in \textbf{bold} while the second is marked with \uline{underline}. $*$ indicates we report results from the corresponding paper. Some work in top block interleaves the natural language and code, which we classify as enhanced P-CoT or PAL \cite{gao2023pal}, reporting P-CoT performance. \textbf{Note} that the SVAMP performance of MathGenie and HTL is evaluated in an OOD setting. }
    \label{tab:sft_ppo_result}
\end{table*}

\section{Experiments}
\subsection{Datasets and Models.}
We conduct experiments on three widely used mathematical reasoning datasets spanning different difficulty levels: SVAMP \cite{Patel_Bhattamishra_Goyal_2021}, GSM8K \cite{Cobbe_Kosaraju_Bavarian_Hilton_Nakano_Hesse_Schulman_2021}, and MathQA \cite{DBLP:conf/naacl/AminiGLKCH19}. For MathQA, we convert the multiple-choice (i.e., ABCD) format into a numeric version to fit the unified input-output form. As for data sources, construction details, and train sizes, please refer to the Appendix \ref{dataset}.

We choose LLaMA2-Base-7B \cite{DBLP:journals/corr/abs-2307-09288} and CodeLLaMA-7B \cite{DBLP:journals/corr/abs-2308-12950} as our foundation models due to their stability and widespread usage. Additionally, compared to LLaMA2, CodeLLaMA includes extra 500B code tokens, which help validate the differing performances of Parrot on code-expert and non-code-expert models. We also conduct experiments on LLaMA-3-8B \cite{grattafiori2024llama}, LLaMA-3.2-3B\footnote{\url{ https://huggingface.co/meta-LlaMA/LlaMA-3.2-3B}}, and Qwen-2.5-1.5B\footnote{\url{https://huggingface.co/Qwen/Qwen2.5-1.5B}} with more complex MathQA to validate the applicability of Parrot in Section \ref{app}.

\subsection{Baselines.}
Our work aims to jointly enhance the performance of P-CoT and N-CoT through mutual promotion, primarily employing hybrid training and reinforcement learning methods. We use the following methods as baselines: 
\paragraph{Standard SFT and RL methods.} SFT quantifies the model's ability to learn from P-CoT and N-CoT demonstrations, validating the intrinsic advantages and drawbacks of these two paradigms. RL trains models by searching and learning \cite{kumar2025llm}, with performance critically dependent on model initialization and reward design. Following \cite{luong2024reft}, we have implemented the Online Self-Learning (On-SL) \cite{hoi2021online} and Proximal Policy Optimization (PPO) algorithms. 
\paragraph{MAmmoth} \cite{DBLP:conf/iclr/YueQZFH00C24} trains the model using hybrid N-CoT and P-CoT rationales. For fair comparison, we re-implemented it on our P-CoT and N-CoT datasets and used different prompts for inference.
\paragraph{HTL} \cite{li2024humans} first generates CoT, which is used to guide the generation of P-CoT, and further uses error assessment-based PPO.
\paragraph{Tora} \cite{gou2023tora} uses natural language reasoning interleaved with program-based tool use.
\paragraph{MathGenie} \cite{lu2024mathgenie} employs solution back-translation to enhance the question diversity.
\paragraph{DotaMath} \cite{li2024dotamath} employs the decomposition of thoughts with code assistance and self-correction for mathematical reasoning.
\paragraph{MARIO} \cite{liao2024mario} introduces a novel math dataset and enhanced with a capability to utilize a Python code interpreter.
\paragraph{Proprietary model.} We also incorporate the closed-source model GPT-4 \cite{DBLP:journals/corr/abs-2303-08774}, which represents the advanced performance in mathematical reasoning.



\subsection{Training details.} 
The specific training and implementation details can be found in Appendix \ref{appendix:details}.

\subsection{Experimental Results}
The main experimental results are presented in Table \ref{tab:sft_ppo_result}. We primarily analyze the model performance on N-CoT reasoning and P-CoT reasoning.
\paragraph{Results on N-CoT reasoning.}
Compared to methods that directly generate N-CoT from problems, Parrot N-CoT refers to P-CoT and its intermediate results, which serve as a simple yet effective process supervision as discussed in section \ref{N-CoT}. Meanwhile, P-CoT's concise reasoning steps enable N-CoT to alleviate the issues of redundant information and logical incoherence. 
Overall, we found the following: 1) Generating N-CoT from P-CoT proves highly effective across all benchmarks, exceeding most of the baselines, and the performance improves with the enhancement of P-CoT's performance, 
which is relatively less challenge. 2) Parrot provides an efficient way for obtaining high N-CoT performance. After Parrot SFT, the model achieves significant improvements comparable to baseline RL. For example, LLaMA2-7B performers better, 12.54 on average, while RL requires considerable resources for searching and learning. 3) The benefit is more pronounced on the challenging
MathQA dataset. While the model can't effectively learning using pure natural language, P-CoT compensates for this limitation.
4) The performance of N-CoT even outperforms P-CoT on LLaMA2-7B for MathQA dataset. We hypothesize this is due to natural language being more suited for semantic analysis and planning of complex problems with clear logic and process signals \cite{li2024humans}. The Parrot N-CoT example is provided in Appendix \ref{gains from parrot training}.
\paragraph{Results on P-CoT reasoning.}
Similarly, the Parrot's average P-CoT performance is on par with or surpasses corresponding baselines, highlighting the significance of information retrieval and transferability afforded by hybrid training. A detailed subtask ablation will be provided in section \ref{sec: sub ab}. In addition, we found that: 1) Compared to the baseline RL, Parrot RL demonstrates clear improvements, with gains of 2.33 and 1.21 on LLaMA2 and CodeLLaMA. This indicates that models with proper initialization and reward design exhibit enhanced exploration capabilities. 2) The model's performance on Parrot On-SL has declined, likely as a result of overfitting stemming from the combination of hybrid training and the absence of negative examples during this phase.

\begin{table*}[t]
\caption{The results of ablation experiments on Parrot subtasks. \textbf{IR.} refers to information retrieval and \textbf{PC. w/o im} is paradigms conversion without intermediate results while \textbf{PC. w/ im} is with intermediate results. 
    }
\small
    \centering
    \renewcommand\arraystretch{1.3}
    \adjustbox{max width=1.0\linewidth}{
    \begin{tabular}{l|l|cccccc|cc}
        \Xhline{1pt}
        \multirow{2}{*}{\bf Model} & \multirow{2}{*}{\bf Subtask} & \multicolumn{2}{c}{\bf GSM8K} & \multicolumn{2}{c}{\bf SVAMP} & \multicolumn{2}{c}{\textbf{MathQA}$_\text{numeric}$} & \multicolumn{2}{c}{\bf Average}\\
        &  & \textbf{N-CoT} & \textbf{P-CoT} & \textbf{N-CoT} & \textbf{P-CoT} & \textbf{N-CoT} & \textbf{P-CoT} & \textbf{N-CoT} & \textbf{P-CoT}  \\
        \hline
\multirow{5}{*}{LLaMA2} & N/P-CoT                                      & $44.05$ & $58.61$ & $58.60$ & $69.50$ & $22.62$ & $46.04$  & $41.76$ & $58.05$ \\ 
& IR. + N/P-CoT                              & $45.19$ & $57.85$ & $58.40$ & $67.10$ & $26.23$ & $46.20$  & $43.27$ & $57.05$ \\
& P-CoT + PC. w/o im                              & $49.43$ & $59.06$ & $59.60$ & $71.60$ & $27.85$ & $45.98$  & $45.63$ & $58.88$ \\
& P-CoT + PC. w/ im                              & $60.81$ & $59.74$ & - & - & $46.82$ & $46.54$  & - & - \\ \cline{2-10}
& Parrot SFT             & $60.81$ & $59.74$ & $59.60$ & $71.60$ & $48.79$ & $46.73$  & $56.40$ & $59.36$ \\
        \hline
\multirow{5}{*}{CodeLLaMA} & N/P-CoT                                      & $44.88$ & $65.05$ & $56.70$ & $75.50$ & $22.37$ & $47.04$  & $41.32$ & $62.53$ \\ 
& IR. + N/P-CoT                              & $45.34$ & $65.96$ & $56.20$ & $75.60$ & $22.55$ & $47.23$  & $41.36$ & $62.93$ \\
& P-CoT + PC. w/o im                              & $50.42$ & $65.13$ & $62.90$ & $77.60$ & $25.17$ & $46.54$  & $46.16$ & $63.09$ \\
& P-CoT + PC. w/ im                              & $64.90$ & $66.19$ & - & - & $42.73$ & $46.86$  & - & - \\ \cline{2-10}
& Parrot SFT             & $64.90$ & $66.19$ & $62.90$ & $77.60$ & $46.84$ & $49.03$  & $58.21$ & $64.27$ \\

        \Xhline{1pt}
    \end{tabular}
    }
    \label{tab:ab res}
\end{table*}

\section{Analysis and Discussion}\label{discussion}
Parrot primarily achieves mutual enhancement through three specially designed subtasks and hybrid training. We give a detail ablation analysis in section \ref{sec: sub ab}, and we further discuss: 1) The impact of the N-CoT penalty in P-CoT PPO in section \ref{pene}, 2) With the aid of P-CoT, which errors are solved and how N-CoT's quality in section \ref{N-CoT}, 3) The applicability of Parrot training pipeline in section \ref{app}.


\begin{table}[ht]
\scriptsize
\centering
\setlength{\belowdisplayskip}{5pt}
\renewcommand{\arraystretch}{1.3}
\begin{tabular}{l|cccc}
    \Xhline{1pt}
    Method & LLaMA2 & CodeLLaMA & Qwen-2.5 & LLaMA-3.2 \\
    \hline
    P-CoT SFT             & 46.04 & 47.04 & 48.29 & 41.56 \\
    IR. + P-CoT            & 46.20 & 47.23 & 48.91 & 41.87 \\
    P-CoT + PC.            & 46.54 & 46.86 & 48.04 & 42.43 \\
    Parrot SFT            & 46.73 & 49.03 & 50.53 & 44.42 \\
    \Xhline{1pt}
\end{tabular}
\caption{The results of \textbf{IR.} ablation experiments. We use Qwen-2.5-1.5B and LLaMA-3.2-3B. Compared with Parrot SFT, P-CoT + PC. omits the IR. subtask. }
\label{tab: IR ablations}
\end{table}

\subsection{Ablations Analysis }\label{sec: sub ab}
\paragraph{Subtask Ablation.} \label{subtask ab}
We analyze each subtask's role sequentially, and the results are in Table \ref{tab:ab res}: 1) For \textbf{Information Retrieval}, which is designed to enable P-CoT's variable definitions. However, we find it has minimal impact on less challenging datasets such as SVAMP and GSM8K, where P-CoT has done well, information retrieval risks misleading for repetitive information, there by we only applied this subtask on MathQA. 

Further, another phenomenon we observe is that although including IR. subtask achieves limited improvements compared to pure P-CoT SFT (47.04 to 47.23 on CodeLLaMA, 46.04 to 46.20 on LLaMA2) on the MathQA dataset, after Parrot training, the stronger model CodeLLaMA gains a notable improvement compared to without IR. (P-CoT + PC. w/ im 46.86 to Parrot SFT 49.03, 2.17). We envision that the model learns to identify whether the key information is accurate and how to utilize it from the transferability of subtask hybrid training, and ultimately generates better quality P-CoT. We also conduct experiments on advanced models Qwen-2.5-1.5B, LLaMA-3.2-3B, and reach similar conclusions. The results can be found in Table \ref{tab: IR ablations}.
2) For \textbf{P-CoT Reasoning}, the results of removing this subtask is IR. + N-CoT for MathQA and N-CoT for GSM8K, SVAMP. Compared to Parrot, this causes a significant performance degradation on LlaMA2 with 22.56 on MathQA, 16.76 on GSM8K, and similar trends on CodeLLaMA. To further explore this notable degradation, we introduce two different settings in the 3) \textbf{Paradigms Conversion} ablation: with and without P-CoT's intermediate results. We are intrigued to find that intermediate results prove essential. The model learns concise reasoning from P-CoT intermediate steps, with the improvements of 11.38, 14.48, and 18.97, 17.56 on GSM8K and MathQA N-CoT for LLaMA2 and CodeLLaMA. 


\paragraph{Hybrid Training.}
Inspired by \cite{DBLP:conf/iclr/YueQZFH00C24}, we apply hybrid training to expect semantic transfer from different linguistic solutions for mutual enhancement, especially on P-CoT with the explicit thinking process \cite{lin2024lean}. Compared to baseline P-CoT, P-CoT + PC. consistently outperforms, achieving +1.1 on GSM8K and +2.1 on SVAMP, but shows a slight (-0.18) decline on MathQA. We hypothesize that while this enhances semantic diversity, it also interferes with P-CoT's precise variable definition. By integrating the information retrieval subtask, Parrot SFT yields a 1.99 improvement over P-CoT SFT on MathQA, validating our hypothesis.





\begin{figure}[t]
    \centering
    \begin{subfigure}[b]{0.225\textwidth}
        \centering
        \includegraphics[width=\textwidth]{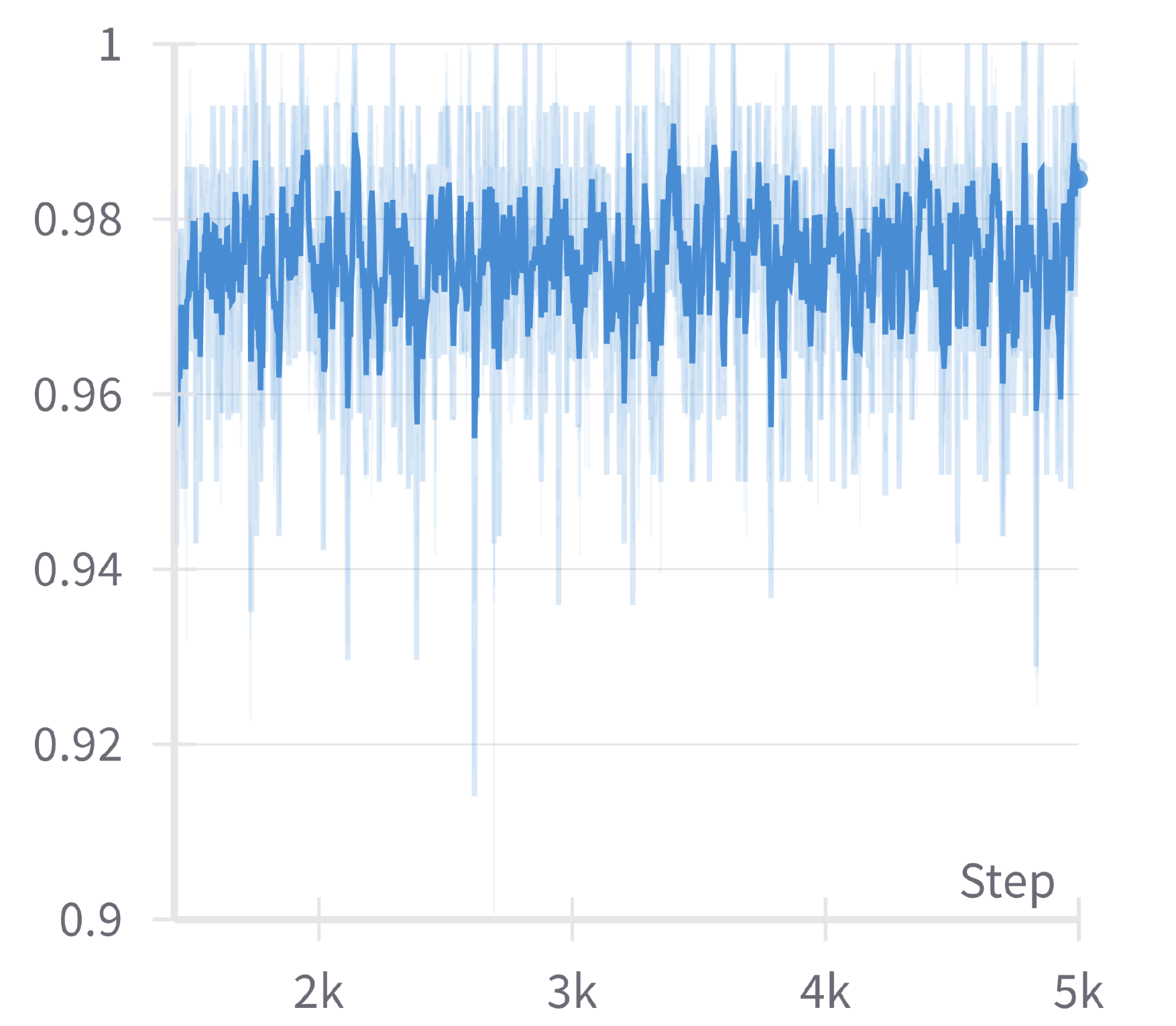}
    \end{subfigure}
    \begin{subfigure}[b]{0.225\textwidth}
        \centering
        \includegraphics[width=\textwidth]{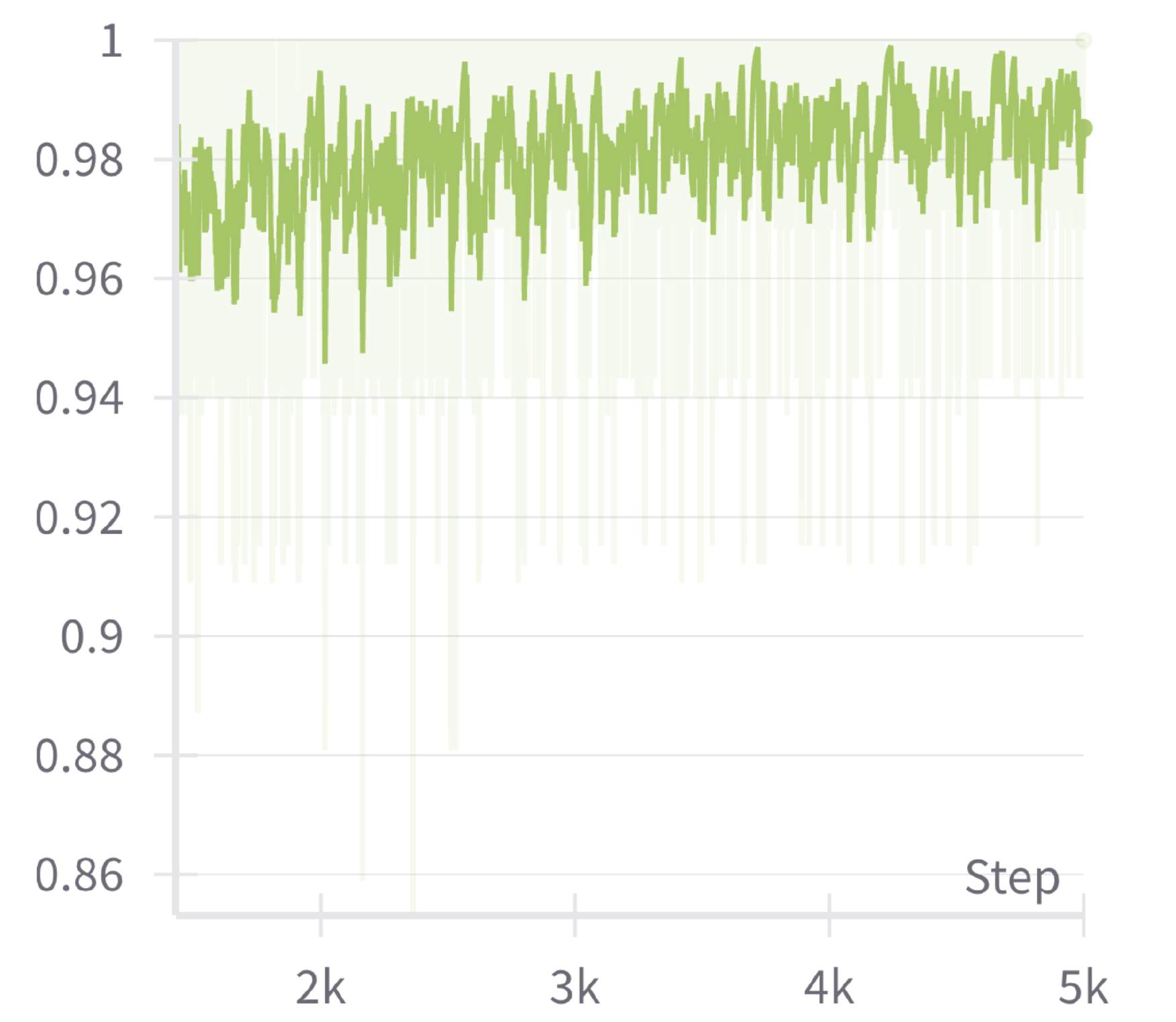}
    \end{subfigure}
    \caption{The training accuracy of LLaMA-2-7B on SVAMP for P-CoT RL. The left figure is without the converted penalty while the right is with the penalty.}
    \label{fig:converted penalty}
\end{figure}

\subsection{The impact of the N-CoT penalty in P-CoT PPO.} \label{pene}
Figure \ref{fig:converted penalty} shows the training accuracy of LLaMA-2-7B on SVAMP for P-CoT PPO. SVAMP is a relatively simple task, allowing the model to produce reasonably good samples and achieve high rewards during the early exploration stages. Without the penalty signal related to P-CoT quality, the model tends to fall into the trap of suboptimal overfitting. In contrast, as shown in the right figure, the model exhibits steady and continuous improvement.

\begin{figure}[t]
    \centering
    \includegraphics[width=1.0\linewidth]{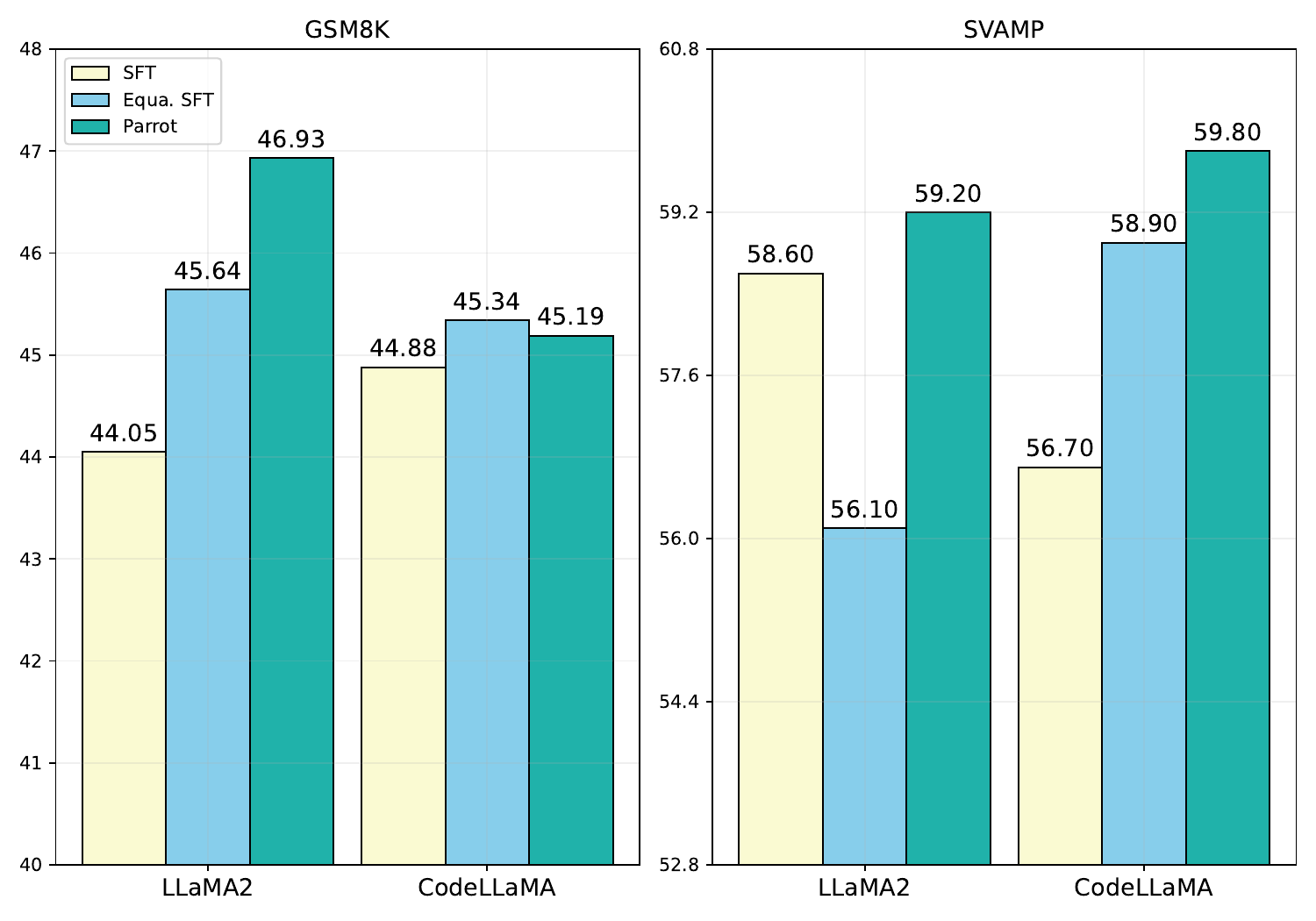}
    \caption{The results of performing SFT using the original N-CoT and the converted N-CoT data. In the left, SFT represents the original data size, while Equ. SFT refers to randomly expanding data to match the scale of the On-SL collections. Parrot denotes the collected data. We collect the correct N-CoT data from 3 epoch Parrot On-SL and perform supervised training after deduplicating.}
    \label{fig:converted data}
\end{figure}

\subsection{Error Reduction and N-CoT Quality Gains.} \label{N-CoT}
\paragraph{P-CoT Alleviates Computational Errors and Logical Inconsistencies in N-CoT.} As presented in section \ref{Pre}, we also statistically analyzed the error types of N-CoT after Parrot SFT. The comparison results are shown in the top of the Figure \ref{fig:error type}. Besides the significant reduction in computational errors (From 578 to 445 on LLaMA2, from 699 to 494 on CodeLLaMA), logical inconsistencies also significantly decreased, particularly on LLaMA2 (From 403 to 171), where P-CoT's intermediate results serve as a simple and effective process signal guiding the N-CoT reasoning. On one hand, we hope to use intermediate results to alleviate the calculation error of N-CoT. On the other hand, the intermediate variables in P-CoT are often linked to context, helping alleviate logical inconsistencies, and the inference process is provided in Appendix \ref{sub prom}. Due to incremental training on the code corpus, the decline in CodeLLaMA is relatively slight.

\paragraph{Better N-CoT training data obtained from P-CoT.} We collect the converted N-CoT during 3 epoch Parrot On-SL training as the model ceases to merely generate high-quality data after several epochs due to the limited efficacy in exploration \cite{DBLP:journals/corr/abs-2404-14387}. We perform SFT and report the best epoch results. The results are in Figure \ref{fig:converted data}. For fairness, we also randomly expand the original data to match the scale of the collected data. We observe two intriguing findings: 1) In the left, data expansion improves the efficacy on GSM8K, resulting in a performance gain of 1.59 on LLaMA2 and 0.46 on CodeLLaMA. However, for SVAMP, performance degrades with reductions of 2.5 on LLaMA2, which may be due to overfitting for its simplicity.
2) The performance of N-CoT obtained from P-CoT consistently exceeds the original data, even with no evidence of overfitting on SVAMP, which further demonstrates that with the aid of P-CoT, the model generates high-quality N-CoT.


\begin{table}[ht]
\centering
\renewcommand\arraystretch{1.3}
\small
\resizebox{0.95\linewidth}{!}{
\begin{tabular}{l|cc}
\toprule
\textbf{Training Methods} & \textbf{N-CoT} & \textbf{P-CoT}  \\
\Xhline{1pt}
LLaMA-3-8B + SFT & 39.13 & 48.54  \\
LLaMA-3-8B + Parrot SFT & 52.03 & 50.28\\
\Xhline{1pt}
LLaMA-3.2-3B + SFT & 31.93 & 41.56  \\
LLaMA-3.2-3B + Parrot SFT & 41.00 & 44.42\\
\Xhline{1pt}
Qwen-2.5-1.5B + SFT & 32.15 & 48.29  \\
Qwen-2.5-1.5B + Parrot SFT & 47.58 & 50.53\\
\Xhline{1pt}
\end{tabular}}
\caption{The Parrot results on the MathQA dataset with three different models.}
\label{tab:LlamMA-3.2}
\end{table}

\subsection{The applicability of Parrot.} \label{app}
We additionally introduce some up-to-date works, MathGenie\cite{lu2024mathgenie}, ToRA \cite{gou2023tora}, DotaMath \cite{li2024dotamath}, MARIO \cite{liao2024mario}. The performance of our model is generally close to or slightly lower than these results (74.53 vs 71.7, 72.6, 79.6, 78.4). Considering that they either used more data or a stronger base model, this gap is relatively acceptable and proves the data efficiency of Parrot.

To verify Parrot’s versatility, we also apply the Parrot pipeline to LLaMA-3-8B, LLaMA-3.2-3B, and Qwen-2.5-1.5B on the MathQA dataset, which is more challenging than GSM8K \cite{luong2024reft} and where we integrate all subtasks.
The consistent improvements across different model sizes and families from Table \ref{tab:LlamMA-3.2} indicate Parrot has the broad applicability, and consistent with previous experiments, the improvement of N-CoT is significant.

\section{Related Work}
\paragraph{Mathematical Reasoning through CoT.}
Significant progress has been made in mathematical reasoning using large language models (LLMs) through chain-of-thought prompting~\citep{wei2022chain} recently.  Specifically,~\citet{fu2022complexity} introduced the concept of complexity-based prompting, demonstrating that LLMs tend to favor long reasoning chains, which often lead to better performance. Recent works such as~\citep{guo2025deepseek} and ~\citep{team2025kimi} have also verified the contribution of long thought chains to reasoning ability. Despite these significant advancements, ensuring the correctness of the chain of thought remains a challenge.  

\paragraph{Design of CoT in Mathematical Reasoning.}
Due to the difficulty in verifying the correctness of the CoT in natural language, a large number of studies have focused on the design of the CoT in mathematical reasoning. The determinacy of programming languages has made the program-assisted method a powerful tool for LLMs to solve mathematical problems.~\citet{chen2022program} has developed a strategy to ensure the consistency of answers between program CoT and natural language CoT, aiming to enhance the reliability of the CoT. Similarly,~\citet{gao2023pal} executes tasks through a Python interpreter to mitigate calculation errors in the natural language CoT. ~\citet{jie2023design} has conducted a comprehensive analysis and comparison of the thought chains in natural language CoT and program CoT, revealing their unique characteristics and potential advantages. 

\paragraph{Exploration of Mathematical Reasoning Paradigms.}
Specifically for solving math problems using LLMs, the main training paradigms revolve around Supervised Fine-Tuning (SFT), Reinforcement Learning (RL), and re-ranking.\citet{DBLP:journals/corr/abs-2211-14275} and \citet{lightman2023let} trained an outcome-based or process-based reward model to perform reranking~\cite{Cobbe_Kosaraju_Bavarian_Hilton_Nakano_Hesse_Schulman_2021}, attaining significantly superior performance compared to the methods of supervised fine-tuning (SFT) and majority voting~\citep{wang2022self}. \citet{luong2024reft} and \citet{guo2025deepseek}further enhanced the generalization ability of LLMs in problem-solving through RL.


\section{Conclusion and Future Work.}
In this paper, we conduct a detailed analysis of error types of P-CoT and N-CoT paradigms and seek to merge the benefits of these two paradigms for mutual promotion, based on which we propose \textbf{Parrot}, a novel training pipeline that integrates three target-designed subtasks for the sequential P-CoT and N-CoT generations. We employ a hybrid training strategy to enhance transferability across pipeline subtasks and analyze the impact of each sub-task in detail. We further expand the pipeline with search and learning algorithms and introduce a converted N-CoT reward to alleviate the sparse issue in the P-CoT RL phase. Extensive experiments demonstrate that Parrot can simultaneously improve both P-CoT and N-CoT performance, especially on N-CoT. In the future, we plan to apply the Parrot pipeline to other reasoning domains such as math proving \cite{lin2024lean}.

\section*{Limitations}

This study has several limitations. First, the proposed sub-task hybrid training strategy demonstrates high sensitivity to data distribution, requiring carefully balanced datasets for optimal performance. Additionally, the resource-intensive search and learning algorithms necessitate substantial computational resources, with model initialization playing a critical yet potentially understudied role in multi-task scenarios. Second, our study focuses solely on mathematical reasoning, leaving other critical reasoning domains (e.g., logical, scientific, and ethical reasoning) unexplored, which limits the broader applicability of our methodology. Furthermore, we did not conduct experiments on the complex MATH dataset \cite{hendrycks2021measuring} for several reasons: 1) Most MATH problems and solutions are written in LaTeX, which highlights the limitations of natural language in key information retrieval and resolution conversion. 2) The limited problems available for both P-CoT and N-CoT made it difficult for models to generate P-CoT. Instead, we conducted experiments on the MathQA dataset, which presents problems in natural language, using the LLaMA3 and Qwen-2.5 series models. Future research could investigate more stable training paradigms and expand the research framework to include additional cognitive tasks, enhancing the robustness and broader applicability of our approach.

\section*{Ethical Considerations}
This research employs closed-source models for data synthesis and fine-tunes open-source models to enhance mathematical reasoning. We adhere to ACL's ethical policies and have rigorously checked the data to mitigate ethical and privacy concerns. Responsible use of LLMs is emphasized to avoid harmful or biased outputs. Reinforcement learning was also employed, which may lead to high resource consumption and environmental impacts.

\section*{Acknowledgements}

The authors wish to thank the anonymous reviewers for their helpful comments. This work was partially funded by the Science and Technology Commission of Shanghai Municipality (No. 24511103100), National Natural Science Foundation of China (No.62476061,62206057), Shanghai Rising-Star Program (23QA1400200), and Natural Science Foundation of Shanghai (23ZR1403500).

\normalem
\bibliography{custom}

\begin{thebibliography}{57}
\providecommand{\natexlab}[1]{#1}

\bibitem[{Amini et~al.(2019)Amini, Gabriel, Lin, Koncel{-}Kedziorski, Choi, and Hajishirzi}]{DBLP:conf/naacl/AminiGLKCH19}
Aida Amini, Saadia Gabriel, Shanchuan Lin, Rik Koncel{-}Kedziorski, Yejin Choi, and Hannaneh Hajishirzi. 2019.
\newblock \href {https://doi.org/10.18653/V1/N19-1245} {Mathqa: Towards interpretable math word problem solving with operation-based formalisms}.
\newblock In \emph{Proceedings of the 2019 Conference of the North American Chapter of the Association for Computational Linguistics: Human Language Technologies, {NAACL-HLT} 2019, Minneapolis, MN, USA, June 2-7, 2019, Volume 1 (Long and Short Papers)}, pages 2357--2367. Association for Computational Linguistics.

\bibitem[{Anthony et~al.(2017)Anthony, Tian, and Barber}]{DBLP:conf/nips/AnthonyTB17}
Thomas Anthony, Zheng Tian, and David Barber. 2017.
\newblock \href {https://proceedings.neurips.cc/paper/2017/hash/d8e1344e27a5b08cdfd5d027d9b8d6de-Abstract.html} {Thinking fast and slow with deep learning and tree search}.
\newblock In \emph{Advances in Neural Information Processing Systems 30: Annual Conference on Neural Information Processing Systems 2017, December 4-9, 2017, Long Beach, CA, {USA}}, pages 5360--5370.

\bibitem[{Chen et~al.(2022)Chen, Ma, Wang, and Cohen}]{chen2022program}
Wenhu Chen, Xueguang Ma, Xinyi Wang, and William~W Cohen. 2022.
\newblock Program of thoughts prompting: Disentangling computation from reasoning for numerical reasoning tasks.
\newblock \emph{arXiv preprint arXiv:2211.12588}.

\bibitem[{Chen et~al.(2025)Chen, He, Xi, Guo, Hong, Zhang, Zheng, Li, Gui, Li et~al.}]{chen2025better}
Wenxiang Chen, Wei He, Zhiheng Xi, Honglin Guo, Boyang Hong, Jiazheng Zhang, Rui Zheng, Nijun Li, Tao Gui, Yun Li, et~al. 2025.
\newblock Better process supervision with bi-directional rewarding signals.
\newblock \emph{arXiv preprint arXiv:2503.04618}.

\bibitem[{Cobbe et~al.(2021)Cobbe, Kosaraju, Bavarian, Hilton, Nakano, Hesse, and Schulman}]{Cobbe_Kosaraju_Bavarian_Hilton_Nakano_Hesse_Schulman_2021}
Karl Cobbe, Vineet Kosaraju, Mohammad Bavarian, Jacob Hilton, Reiichiro Nakano, Christopher Hesse, and John Schulman. 2021.
\newblock Training verifiers to solve math word problems.
\newblock \emph{Cornell University - arXiv,Cornell University - arXiv}.

\bibitem[{Fu et~al.(2022)Fu, Peng, Sabharwal, Clark, and Khot}]{fu2022complexity}
Yao Fu, Hao Peng, Ashish Sabharwal, Peter Clark, and Tushar Khot. 2022.
\newblock Complexity-based prompting for multi-step reasoning.
\newblock In \emph{The Eleventh International Conference on Learning Representations}.

\bibitem[{Gao et~al.(2023)Gao, Madaan, Zhou, Alon, Liu, Yang, Callan, and Neubig}]{gao2023pal}
Luyu Gao, Aman Madaan, Shuyan Zhou, Uri Alon, Pengfei Liu, Yiming Yang, Jamie Callan, and Graham Neubig. 2023.
\newblock Pal: Program-aided language models.
\newblock In \emph{International Conference on Machine Learning}, pages 10764--10799. PMLR.

\bibitem[{Gou et~al.(2023)Gou, Shao, Gong, Shen, Yang, Huang, Duan, and Chen}]{gou2023tora}
Zhibin Gou, Zhihong Shao, Yeyun Gong, Yelong Shen, Yujiu Yang, Minlie Huang, Nan Duan, and Weizhu Chen. 2023.
\newblock Tora: A tool-integrated reasoning agent for mathematical problem solving.
\newblock \emph{arXiv preprint arXiv:2309.17452}.

\bibitem[{Grattafiori et~al.(2024)Grattafiori, Dubey, Jauhri, Pandey, Kadian, Al-Dahle, Letman, Mathur, Schelten, Vaughan et~al.}]{grattafiori2024llama}
Aaron Grattafiori, Abhimanyu Dubey, Abhinav Jauhri, Abhinav Pandey, Abhishek Kadian, Ahmad Al-Dahle, Aiesha Letman, Akhil Mathur, Alan Schelten, Alex Vaughan, et~al. 2024.
\newblock The llama 3 herd of models.
\newblock \emph{arXiv preprint arXiv:2407.21783}.

\bibitem[{Gugger et~al.(2022)Gugger, Debut, Wolf, Schmid, Mueller, Mangrulkar, Sun, and Bossan}]{accelerate}
Sylvain Gugger, Lysandre Debut, Thomas Wolf, Philipp Schmid, Zachary Mueller, Sourab Mangrulkar, Marc Sun, and Benjamin Bossan. 2022.
\newblock Accelerate: Training and inference at scale made simple, efficient and adaptable.
\newblock \url{https://github.com/huggingface/accelerate}.

\bibitem[{Guo et~al.(2025)Guo, Yang, Zhang, Song, Zhang, Xu, Zhu, Ma, Wang, Bi et~al.}]{guo2025deepseek}
Daya Guo, Dejian Yang, Haowei Zhang, Junxiao Song, Ruoyu Zhang, Runxin Xu, Qihao Zhu, Shirong Ma, Peiyi Wang, Xiao Bi, et~al. 2025.
\newblock Deepseek-r1: Incentivizing reasoning capability in llms via reinforcement learning.
\newblock \emph{arXiv preprint arXiv:2501.12948}.

\bibitem[{Hendrycks et~al.(2021)Hendrycks, Burns, Kadavath, Arora, Basart, Tang, Song, and Steinhardt}]{hendrycks2021measuring}
Dan Hendrycks, Collin Burns, Saurav Kadavath, Akul Arora, Steven Basart, Eric Tang, Dawn Song, and Jacob Steinhardt. 2021.
\newblock Measuring mathematical problem solving with the math dataset.
\newblock \emph{arXiv preprint arXiv:2103.03874}.

\bibitem[{Hoi et~al.(2021)Hoi, Sahoo, Lu, and Zhao}]{hoi2021online}
Steven~CH Hoi, Doyen Sahoo, Jing Lu, and Peilin Zhao. 2021.
\newblock Online learning: A comprehensive survey.
\newblock \emph{Neurocomputing}, 459:249--289.

\bibitem[{Jie et~al.(2023)Jie, Luong, Zhang, Jin, and Li}]{jie2023design}
Zhanming Jie, Trung~Quoc Luong, Xinbo Zhang, Xiaoran Jin, and Hang Li. 2023.
\newblock Design of chain-of-thought in math problem solving.
\newblock \emph{arXiv preprint arXiv:2309.11054}.

\bibitem[{Kazemi et~al.(2012)Kazemi, Yektayar, and Abad}]{kazemi2012investigation}
Farhad Kazemi, Mozafar Yektayar, and Ali Mohammadi~Bolban Abad. 2012.
\newblock Investigation the impact of chess play on developing meta-cognitive ability and math problem-solving power of students at different levels of education.
\newblock \emph{Procedia-Social and Behavioral Sciences}, 32:372--379.

\bibitem[{Krawec(2014)}]{krawec2014problem}
Jennifer~L Krawec. 2014.
\newblock Problem representation and mathematical problem solving of students of varying math ability.
\newblock \emph{Journal of Learning Disabilities}, 47(2):103--115.

\bibitem[{Kullback and Leibler(1951)}]{kullback1951information}
Solomon Kullback and Richard~A Leibler. 1951.
\newblock On information and sufficiency.
\newblock \emph{The annals of mathematical statistics}, 22(1):79--86.

\bibitem[{Kumar et~al.(2024)Kumar, Zhuang, Agarwal, Su, Co{-}Reyes, Singh, Baumli, Iqbal, Bishop, Roelofs, Zhang, McKinney, Shrivastava, Paduraru, Tucker, Precup, Behbahani, and Faust}]{DBLP:journals/corr/abs-2409-12917}
Aviral Kumar, Vincent Zhuang, Rishabh Agarwal, Yi~Su, John~D. Co{-}Reyes, Avi Singh, Kate Baumli, Shariq Iqbal, Colton Bishop, Rebecca Roelofs, Lei~M. Zhang, Kay McKinney, Disha Shrivastava, Cosmin Paduraru, George Tucker, Doina Precup, Feryal M.~P. Behbahani, and Aleksandra Faust. 2024.
\newblock \href {https://doi.org/10.48550/ARXIV.2409.12917} {Training language models to self-correct via reinforcement learning}.
\newblock \emph{CoRR}, abs/2409.12917.

\bibitem[{Kumar et~al.(2025)Kumar, Ashraf, Thawakar, Anwer, Cholakkal, Shah, Yang, Torr, Khan, and Khan}]{kumar2025llm}
Komal Kumar, Tajamul Ashraf, Omkar Thawakar, Rao~Muhammad Anwer, Hisham Cholakkal, Mubarak Shah, Ming-Hsuan Yang, Phillip~HS Torr, Fahad~Shahbaz Khan, and Salman Khan. 2025.
\newblock Llm post-training: A deep dive into reasoning large language models.
\newblock \emph{arXiv preprint arXiv:2502.21321}.

\bibitem[{Le et~al.(2022)Le, Wang, Gotmare, Savarese, and Hoi}]{le2022coderl}
Hung Le, Yue Wang, Akhilesh~Deepak Gotmare, Silvio Savarese, and Steven Chu~Hong Hoi. 2022.
\newblock Coderl: Mastering code generation through pretrained models and deep reinforcement learning.
\newblock \emph{Advances in Neural Information Processing Systems}, 35:21314--21328.

\bibitem[{Li et~al.(2024{\natexlab{a}})Li, Dong, Xue, Peng, Wang, and Liu}]{li2024dotamath}
Chengpeng Li, Guanting Dong, Mingfeng Xue, Ru~Peng, Xiang Wang, and Dayiheng Liu. 2024{\natexlab{a}}.
\newblock Dotamath: Decomposition of thought with code assistance and self-correction for mathematical reasoning.
\newblock \emph{arXiv preprint arXiv:2407.04078}.

\bibitem[{Li et~al.(2024{\natexlab{b}})Li, He, Wang, Wang, and He}]{li2024humans}
Long Li, Xuzheng He, Haozhe Wang, Linlin Wang, and Liang He. 2024{\natexlab{b}}.
\newblock How do humans write code? large models do it the same way too.
\newblock \emph{arXiv preprint arXiv:2402.15729}.

\bibitem[{Li et~al.(2024{\natexlab{c}})Li, Wang, Li, Guo, Zhang, and Feng}]{li2024evaluating}
Xiaoyuan Li, Wenjie Wang, Moxin Li, Junrong Guo, Yang Zhang, and Fuli Feng. 2024{\natexlab{c}}.
\newblock Evaluating mathematical reasoning of large language models: A focus on error identification and correction.
\newblock \emph{arXiv preprint arXiv:2406.00755}.

\bibitem[{Liang et~al.(2024)Liang, Liu, Niu, Zhang, Zhou, and Yavuz}]{liang2024improving}
Zhenwen Liang, Ye~Liu, Tong Niu, Xiangliang Zhang, Yingbo Zhou, and Semih Yavuz. 2024.
\newblock Improving llm reasoning through scaling inference computation with collaborative verification.
\newblock \emph{arXiv preprint arXiv:2410.05318}.

\bibitem[{Liao et~al.(2024)Liao, Luo, Li, Wu, and Fan}]{liao2024mario}
Minpeng Liao, Wei Luo, Chengxi Li, Jing Wu, and Kai Fan. 2024.
\newblock Mario: Math reasoning with code interpreter output--a reproducible pipeline.
\newblock \emph{arXiv preprint arXiv:2401.08190}.

\bibitem[{Lightman et~al.(2023)Lightman, Kosaraju, Burda, Edwards, Baker, Lee, Leike, Schulman, Sutskever, and Cobbe}]{lightman2023let}
Hunter Lightman, Vineet Kosaraju, Yura Burda, Harri Edwards, Bowen Baker, Teddy Lee, Jan Leike, John Schulman, Ilya Sutskever, and Karl Cobbe. 2023.
\newblock Let's verify step by step.
\newblock \emph{arXiv preprint arXiv:2305.20050}.

\bibitem[{Lin et~al.(2024)Lin, Sun, Welleck, and Yang}]{lin2024lean}
Haohan Lin, Zhiqing Sun, Sean Welleck, and Yiming Yang. 2024.
\newblock Lean-star: Learning to interleave thinking and proving.
\newblock \emph{arXiv preprint arXiv:2407.10040}.

\bibitem[{Loshchilov and Hutter(2019)}]{DBLP:conf/iclr/LoshchilovH19}
Ilya Loshchilov and Frank Hutter. 2019.
\newblock \href {https://openreview.net/forum?id=Bkg6RiCqY7} {Decoupled weight decay regularization}.
\newblock In \emph{7th International Conference on Learning Representations, {ICLR} 2019, New Orleans, LA, USA, May 6-9, 2019}. OpenReview.net.

\bibitem[{Lu et~al.(2024{\natexlab{a}})Lu, Zhou, Ren, Wang, Shi, Pan, Zhan, and Li}]{lu2024mathgenie}
Zimu Lu, Aojun Zhou, Houxing Ren, Ke~Wang, Weikang Shi, Junting Pan, Mingjie Zhan, and Hongsheng Li. 2024{\natexlab{a}}.
\newblock Mathgenie: Generating synthetic data with question back-translation for enhancing mathematical reasoning of llms.
\newblock \emph{arXiv preprint arXiv:2402.16352}.

\bibitem[{Lu et~al.(2024{\natexlab{b}})Lu, Zhou, Wang, Ren, Shi, Pan, Zhan, and Li}]{lu2024mathcoder2}
Zimu Lu, Aojun Zhou, Ke~Wang, Houxing Ren, Weikang Shi, Junting Pan, Mingjie Zhan, and Hongsheng Li. 2024{\natexlab{b}}.
\newblock Mathcoder2: Better math reasoning from continued pretraining on model-translated mathematical code.
\newblock \emph{arXiv preprint arXiv:2410.08196}.

\bibitem[{Luo et~al.(2024)Luo, Liu, Liu, Phatale, Guo, Lara, Li, Shu, Zhu, Meng et~al.}]{luo2024improve}
Liangchen Luo, Yinxiao Liu, Rosanne Liu, Samrat Phatale, Meiqi Guo, Harsh Lara, Yunxuan Li, Lei Shu, Yun Zhu, Lei Meng, et~al. 2024.
\newblock Improve mathematical reasoning in language models by automated process supervision.
\newblock \emph{arXiv preprint arXiv:2406.06592}.

\bibitem[{Luong et~al.(2024)Luong, Zhang, Jie, Sun, Jin, and Li}]{luong2024reft}
Trung~Quoc Luong, Xinbo Zhang, Zhanming Jie, Peng Sun, Xiaoran Jin, and Hang Li. 2024.
\newblock Reft: Reasoning with reinforced fine-tuning.
\newblock \emph{arXiv preprint arXiv:2401.08967}.

\bibitem[{OpenAI(2023)}]{DBLP:journals/corr/abs-2303-08774}
OpenAI. 2023.
\newblock \href {https://doi.org/10.48550/ARXIV.2303.08774} {{GPT-4} technical report}.
\newblock \emph{CoRR}, abs/2303.08774.

\bibitem[{Patel et~al.(2021)Patel, Bhattamishra, and Goyal}]{Patel_Bhattamishra_Goyal_2021}
Arkil Patel, Satwik Bhattamishra, and Navin Goyal. 2021.
\newblock \href {https://doi.org/10.18653/v1/2021.naacl-main.168} {Are nlp models really able to solve simple math word problems?}
\newblock In \emph{Proceedings of the 2021 Conference of the North American Chapter of the Association for Computational Linguistics: Human Language Technologies}.

\bibitem[{Rajbhandari et~al.(2020)Rajbhandari, Rasley, Ruwase, and He}]{9355301}
Samyam Rajbhandari, Jeff Rasley, Olatunji Ruwase, and Yuxiong He. 2020.
\newblock \href {https://doi.org/10.1109/SC41405.2020.00024} {Zero: Memory optimizations toward training trillion parameter models}.
\newblock In \emph{SC20: International Conference for High Performance Computing, Networking, Storage and Analysis}, pages 1--16.

\bibitem[{Rasley et~al.(2020)Rasley, Rajbhandari, Ruwase, and He}]{rasley2020deepspeed}
Jeff Rasley, Samyam Rajbhandari, Olatunji Ruwase, and Yuxiong He. 2020.
\newblock Deepspeed: System optimizations enable training deep learning models with over 100 billion parameters.
\newblock In \emph{Proceedings of the 26th ACM SIGKDD International Conference on Knowledge Discovery \& Data Mining}, pages 3505--3506.

\bibitem[{Renze and Guven(2024)}]{DBLP:journals/corr/abs-2405-06682}
Matthew Renze and Erhan Guven. 2024.
\newblock \href {https://doi.org/10.48550/ARXIV.2405.06682} {Self-reflection in {LLM} agents: Effects on problem-solving performance}.
\newblock \emph{CoRR}, abs/2405.06682.

\bibitem[{Rozi{\`{e}}re et~al.(2023)Rozi{\`{e}}re, Gehring, Gloeckle, Sootla, Gat, Tan, Adi, Liu, Remez, Rapin, Kozhevnikov, Evtimov, Bitton, Bhatt, Canton{-}Ferrer, Grattafiori, Xiong, D{\'{e}}fossez, Copet, Azhar, Touvron, Martin, Usunier, Scialom, and Synnaeve}]{DBLP:journals/corr/abs-2308-12950}
Baptiste Rozi{\`{e}}re, Jonas Gehring, Fabian Gloeckle, Sten Sootla, Itai Gat, Xiaoqing~Ellen Tan, Yossi Adi, Jingyu Liu, Tal Remez, J{\'{e}}r{\'{e}}my Rapin, Artyom Kozhevnikov, Ivan Evtimov, Joanna Bitton, Manish Bhatt, Cristian Canton{-}Ferrer, Aaron Grattafiori, Wenhan Xiong, Alexandre D{\'{e}}fossez, Jade Copet, Faisal Azhar, Hugo Touvron, Louis Martin, Nicolas Usunier, Thomas Scialom, and Gabriel Synnaeve. 2023.
\newblock \href {https://doi.org/10.48550/ARXIV.2308.12950} {Code llama: Open foundation models for code}.
\newblock \emph{CoRR}, abs/2308.12950.

\bibitem[{Schulman et~al.(2017)Schulman, Wolski, Dhariwal, Radford, and Klimov}]{schulman2017proximal}
John Schulman, Filip Wolski, Prafulla Dhariwal, Alec Radford, and Oleg Klimov. 2017.
\newblock Proximal policy optimization algorithms.
\newblock \emph{arXiv preprint arXiv:1707.06347}.

\bibitem[{Shao et~al.(2024)Shao, Wang, Zhu, Xu, Song, Zhang, Li, Wu, and Guo}]{DBLP:journals/corr/abs-2402-03300}
Zhihong Shao, Peiyi Wang, Qihao Zhu, Runxin Xu, Junxiao Song, Mingchuan Zhang, Y.~K. Li, Y.~Wu, and Daya Guo. 2024.
\newblock \href {https://doi.org/10.48550/ARXIV.2402.03300} {Deepseekmath: Pushing the limits of mathematical reasoning in open language models}.
\newblock \emph{CoRR}, abs/2402.03300.

\bibitem[{Su et~al.(2024)Su, Zhang, Qu, Zhu, Li, Sun, Li, Zhang, and Cheng}]{DBLP:conf/nips/SuZQ0LSLZC24}
Zhaochen Su, Jun Zhang, Xiaoye Qu, Tong Zhu, Yanshu Li, Jiashuo Sun, Juntao Li, Min Zhang, and Yu~Cheng. 2024.
\newblock \href {http://papers.nips.cc/paper\_files/paper/2024/hash/baf4b960d118f838ad0b2c08247a9ebe-Abstract-Datasets\_and\_Benchmarks\_Track.html} {Conflictbank: {A} benchmark for evaluating the influence of knowledge conflicts in llms}.
\newblock In \emph{Advances in Neural Information Processing Systems 38: Annual Conference on Neural Information Processing Systems 2024, NeurIPS 2024, Vancouver, BC, Canada, December 10 - 15, 2024}.

\bibitem[{Tao et~al.(2024)Tao, Lin, Chen, Li, Wu, Li, Jin, Huang, Tao, and Zhou}]{DBLP:journals/corr/abs-2404-14387}
Zhengwei Tao, Ting{-}En Lin, Xiancai Chen, Hangyu Li, Yuchuan Wu, Yongbin Li, Zhi Jin, Fei Huang, Dacheng Tao, and Jingren Zhou. 2024.
\newblock \href {https://doi.org/10.48550/ARXIV.2404.14387} {A survey on self-evolution of large language models}.
\newblock \emph{CoRR}, abs/2404.14387.

\bibitem[{Team et~al.(2025)Team, Du, Gao, Xing, Jiang, Chen, Li, Xiao, Du, Liao et~al.}]{team2025kimi}
Kimi Team, Angang Du, Bofei Gao, Bowei Xing, Changjiu Jiang, Cheng Chen, Cheng Li, Chenjun Xiao, Chenzhuang Du, Chonghua Liao, et~al. 2025.
\newblock Kimi k1. 5: Scaling reinforcement learning with llms.
\newblock \emph{arXiv preprint arXiv:2501.12599}.

\bibitem[{Touvron et~al.(2023)Touvron, Martin, Stone, Albert, Almahairi, Babaei, Bashlykov, Batra, Bhargava, Bhosale, Bikel, Blecher, Canton{-}Ferrer, Chen, Cucurull, Esiobu, Fernandes, Fu, Fu, Fuller, Gao, Goswami, Goyal, Hartshorn, Hosseini, Hou, Inan, Kardas, Kerkez, Khabsa, Kloumann, Korenev, Koura, Lachaux, Lavril, Lee, Liskovich, Lu, Mao, Martinet, Mihaylov, Mishra, Molybog, Nie, Poulton, Reizenstein, Rungta, Saladi, Schelten, Silva, Smith, Subramanian, Tan, Tang, Taylor, Williams, Kuan, Xu, Yan, Zarov, Zhang, Fan, Kambadur, Narang, Rodriguez, Stojnic, Edunov, and Scialom}]{DBLP:journals/corr/abs-2307-09288}
Hugo Touvron, Louis Martin, Kevin Stone, Peter Albert, Amjad Almahairi, Yasmine Babaei, Nikolay Bashlykov, Soumya Batra, Prajjwal Bhargava, Shruti Bhosale, Dan Bikel, Lukas Blecher, Cristian Canton{-}Ferrer, Moya Chen, Guillem Cucurull, David Esiobu, Jude Fernandes, Jeremy Fu, Wenyin Fu, Brian Fuller, Cynthia Gao, Vedanuj Goswami, Naman Goyal, Anthony Hartshorn, Saghar Hosseini, Rui Hou, Hakan Inan, Marcin Kardas, Viktor Kerkez, Madian Khabsa, Isabel Kloumann, Artem Korenev, Punit~Singh Koura, Marie{-}Anne Lachaux, Thibaut Lavril, Jenya Lee, Diana Liskovich, Yinghai Lu, Yuning Mao, Xavier Martinet, Todor Mihaylov, Pushkar Mishra, Igor Molybog, Yixin Nie, Andrew Poulton, Jeremy Reizenstein, Rashi Rungta, Kalyan Saladi, Alan Schelten, Ruan Silva, Eric~Michael Smith, Ranjan Subramanian, Xiaoqing~Ellen Tan, Binh Tang, Ross Taylor, Adina Williams, Jian~Xiang Kuan, Puxin Xu, Zheng Yan, Iliyan Zarov, Yuchen Zhang, Angela Fan, Melanie Kambadur, Sharan Narang, Aur{\'{e}}lien Rodriguez, Robert Stojnic, Sergey Edunov,
  and Thomas Scialom. 2023.
\newblock \href {https://doi.org/10.48550/ARXIV.2307.09288} {Llama 2: Open foundation and fine-tuned chat models}.
\newblock \emph{CoRR}, abs/2307.09288.

\bibitem[{Uesato et~al.(2022)Uesato, Kushman, Kumar, Song, Siegel, Wang, Creswell, Irving, and Higgins}]{DBLP:journals/corr/abs-2211-14275}
Jonathan Uesato, Nate Kushman, Ramana Kumar, H.~Francis Song, Noah~Y. Siegel, Lisa Wang, Antonia Creswell, Geoffrey Irving, and Irina Higgins. 2022.
\newblock \href {https://doi.org/10.48550/ARXIV.2211.14275} {Solving math word problems with process- and outcome-based feedback}.
\newblock \emph{CoRR}, abs/2211.14275.

\bibitem[{Wan et~al.(2024)Wan, Feng, Wen, McAleer, Wen, Zhang, and Wang}]{DBLP:conf/icml/WanFWM00024}
Ziyu Wan, Xidong Feng, Muning Wen, Stephen~Marcus McAleer, Ying Wen, Weinan Zhang, and Jun Wang. 2024.
\newblock \href {https://openreview.net/forum?id=C4OpREezgj} {Alphazero-like tree-search can guide large language model decoding and training}.
\newblock In \emph{Forty-first International Conference on Machine Learning, {ICML} 2024, Vienna, Austria, July 21-27, 2024}. OpenReview.net.

\bibitem[{Wang et~al.(2024)Wang, Li, Shao, Xu, Dai, Li, Chen, Wu, and Sui}]{DBLP:conf/acl/WangLSXDLCWS24}
Peiyi Wang, Lei Li, Zhihong Shao, Runxin Xu, Damai Dai, Yifei Li, Deli Chen, Yu~Wu, and Zhifang Sui. 2024.
\newblock \href {https://doi.org/10.18653/V1/2024.ACL-LONG.510} {Math-shepherd: Verify and reinforce llms step-by-step without human annotations}.
\newblock In \emph{Proceedings of the 62nd Annual Meeting of the Association for Computational Linguistics (Volume 1: Long Papers), {ACL} 2024, Bangkok, Thailand, August 11-16, 2024}, pages 9426--9439. Association for Computational Linguistics.

\bibitem[{Wang et~al.(2022)Wang, Wei, Schuurmans, Le, Chi, Narang, Chowdhery, and Zhou}]{wang2022self}
Xuezhi Wang, Jason Wei, Dale Schuurmans, Quoc Le, Ed~Chi, Sharan Narang, Aakanksha Chowdhery, and Denny Zhou. 2022.
\newblock Self-consistency improves chain of thought reasoning in language models.
\newblock \emph{arXiv preprint arXiv:2203.11171}.

\bibitem[{Wei et~al.(2022)Wei, Wang, Schuurmans, Bosma, Xia, Chi, Le, Zhou et~al.}]{wei2022chain}
Jason Wei, Xuezhi Wang, Dale Schuurmans, Maarten Bosma, Fei Xia, Ed~Chi, Quoc~V Le, Denny Zhou, et~al. 2022.
\newblock Chain-of-thought prompting elicits reasoning in large language models.
\newblock \emph{Advances in neural information processing systems}, 35:24824--24837.

\bibitem[{Xi et~al.(2024)Xi, Chen, Hong, Jin, Zheng, He, Ding, Liu, Guo, Wang et~al.}]{xi2024training}
Zhiheng Xi, Wenxiang Chen, Boyang Hong, Senjie Jin, Rui Zheng, Wei He, Yiwen Ding, Shichun Liu, Xin Guo, Junzhe Wang, et~al. 2024.
\newblock Training large language models for reasoning through reverse curriculum reinforcement learning.
\newblock \emph{arXiv preprint arXiv:2402.05808}.

\bibitem[{Xi et~al.(2023)Xi, Jin, Zhou, Zheng, Gao, Liu, Gui, Zhang, and Huang}]{DBLP:conf/emnlp/XiJZZGLGZH23}
Zhiheng Xi, Senjie Jin, Yuhao Zhou, Rui Zheng, Songyang Gao, Jia Liu, Tao Gui, Qi~Zhang, and Xuanjing Huang. 2023.
\newblock \href {https://doi.org/10.18653/V1/2023.FINDINGS-EMNLP.762} {Self-polish: Enhance reasoning in large language models via problem refinement}.
\newblock In \emph{Findings of the Association for Computational Linguistics: {EMNLP} 2023, Singapore, December 6-10, 2023}, pages 11383--11406. Association for Computational Linguistics.

\bibitem[{Xu et~al.(2024{\natexlab{a}})Xu, Kim, Sharaf, and Awadalla}]{DBLP:conf/iclr/Xu0SA24}
Haoran Xu, Young~Jin Kim, Amr Sharaf, and Hany~Hassan Awadalla. 2024{\natexlab{a}}.
\newblock \href {https://openreview.net/forum?id=farT6XXntP} {A paradigm shift in machine translation: Boosting translation performance of large language models}.
\newblock In \emph{The Twelfth International Conference on Learning Representations, {ICLR} 2024, Vienna, Austria, May 7-11, 2024}. OpenReview.net.

\bibitem[{Xu et~al.(2024{\natexlab{b}})Xu, Qi, Guo, Wang, Wang, Zhang, and Xu}]{DBLP:conf/emnlp/XuQGW0ZX24}
Rongwu Xu, Zehan Qi, Zhijiang Guo, Cunxiang Wang, Hongru Wang, Yue Zhang, and Wei Xu. 2024{\natexlab{b}}.
\newblock \href {https://aclanthology.org/2024.emnlp-main.486} {Knowledge conflicts for llms: {A} survey}.
\newblock In \emph{Proceedings of the 2024 Conference on Empirical Methods in Natural Language Processing, {EMNLP} 2024, Miami, FL, USA, November 12-16, 2024}, pages 8541--8565. Association for Computational Linguistics.

\bibitem[{Yu et~al.(2023)Yu, Gao, and Wang}]{yu2023outcome}
Fei Yu, Anningzhe Gao, and Benyou Wang. 2023.
\newblock Outcome-supervised verifiers for planning in mathematical reasoning.
\newblock \emph{arXiv preprint arXiv:2311.09724}.

\bibitem[{Yue et~al.(2024)Yue, Qu, Zhang, Fu, Huang, Sun, Su, and Chen}]{DBLP:conf/iclr/YueQZFH00C24}
Xiang Yue, Xingwei Qu, Ge~Zhang, Yao Fu, Wenhao Huang, Huan Sun, Yu~Su, and Wenhu Chen. 2024.
\newblock \href {https://openreview.net/forum?id=yLClGs770I} {Mammoth: Building math generalist models through hybrid instruction tuning}.
\newblock In \emph{The Twelfth International Conference on Learning Representations, {ICLR} 2024, Vienna, Austria, May 7-11, 2024}. OpenReview.net.

\bibitem[{Zhang and Yang(2021)}]{zhang2021survey}
Yu~Zhang and Qiang Yang. 2021.
\newblock A survey on multi-task learning.
\newblock \emph{IEEE transactions on knowledge and data engineering}, 34(12):5586--5609.

\bibitem[{Zhong et~al.(2017)Zhong, Xiong, and Socher}]{DBLP:journals/corr/abs-1709-00103}
Victor Zhong, Caiming Xiong, and Richard Socher. 2017.
\newblock \href {https://arxiv.org/abs/1709.00103} {Seq2sql: Generating structured queries from natural language using reinforcement learning}.
\newblock \emph{CoRR}, abs/1709.00103.

\end{thebibliography}

\appendix
\newpage
\section{Preliminary Errors and Prompts}\label{appendix:prompts}
\subsection{Preliminary Error Identifications}\label{preliminary Identifications}
We randomly sampled 50 error cases from each paradigm and the empirical identifications of error types are as follows:
\paragraph{N-CoT Error Identifications:}
\begin{itemize}
\item \small \textbf{Comprehension Error}: Misunderstanding of the problem, omission of conditions.
\item \small \textbf{Logical Inconsistency}: Logic inconsistency between pre-and-post during the reasoning process.
\item \small \textbf{Redundant and Repetitive}: Unnecessary information or overlapping functions, whereas repetitiveness refers to patterns that add little or no substantial value.
\item \small \textbf{Calculation Error}: Basic arithmetic errors and improper application of formulas.
\item \small \textbf{Other Errors}: Other reasoning errors fall outside the scope of the above.
\end{itemize}

\paragraph{P-CoT Error Identifications:}
\begin{itemize}
\item \small \textbf{Comprehension Error}: Misunderstanding of the problem, omission of conditions.
\item \small \textbf{Reasoning Error}: Inadequate reasoning, causal inversion, and circular arguments.
\item \small \textbf{Variable Error}: Incorrect definition and assignment of variables.
\item \small \textbf{Expression Error}: Violations of mathematical operation rules and non-standard Python output.
\end{itemize}

\subsection{Error Evaluation Prompt}\label{Error Evaluation Prompt}
Based on manually verified error types, we use GPT-4 \cite{DBLP:journals/corr/abs-2303-08774} to identify errors in the model's N/P-CoT rationales. The system prompt provided to GPT-4 is:

\begin{tcolorbox}[
    title=GPT-4 Evaluation System Prompt,
    boxrule=0.8pt,
    top=3mm,
    bottom=3mm,
    left=2mm,
    right=2mm,
    beforeafter skip=5mm]

\textbf{\small System Prompt:}
\small
You are a helpful assistant. Analyze the following answer reasoning process, identify the major error in it. \\
Types of errors: \{N/P-CoT Error Identification Types\}. \\
Please analyze the major type of error that may occur. Don't output the explanation. Output the error type directly in the format: \textbf{The error type is: \{\}}. \\
User: \{\textbf{Question}\}'s ground truth answer is \{\textbf{Answer Value}\}. \\
Answer reasoning: \{\textbf{Answer Reasoning Process}\}. 
\end{tcolorbox}

The details of \textbf{N/P-CoT Error Identification Types} can be found in \ref{Pre}, while the \textbf{Answer Reasoning Process} refers to the model's inference outputs.

\subsection{Subtask Prompts and Inference Process.} \label{sub prom}
In this section, we provide the prompts we used and the inference process details of key information, P-CoT, and N-CoT, which align with the three subtasks.

\paragraph{Subtask Prompts:} \label{Subtask Prompts:}
\begin{itemize}
\item \small \textbf{System Prompt}: Question:
\item \small  \textbf{Information Retrieval Prompt}: Answer reasoning: To solve this question, we first find all the key information in the question:
\item \small \textbf{P-CoT Reasoning Prompt}: Please refer to the key information to complete the Python-style solution:
\item \small  \textbf{Paradigm Conversion Prompt}: Please refer to the Python code style solution and the intermediate outputs to complete the natural language style solution. Therefore, the natural language style solution is:
\end{itemize}

\paragraph{Subtask Subtasks:} \label{Subtask Subtasks} 
We divide Parrot Pipeline into three sub-tasks: \textbf{Information Retrieval subtask}, \textbf{P-CoT Reasoning subtask}, \textbf{Paradigm Conversion subtask}.

\begin{tcolorbox}[
    title=Information Retrieval subtask,
    boxrule=0.8pt,
    top=3mm,
    bottom=3mm,
    left=2mm,
    right=2mm,
    beforeafter skip=5mm]

\textbf{\small The input is:} \\
\small
\{System Prompt\} \\
Question \\
\{Information Retrieval Prompt\} \\ 
\\
\textbf{\small The output is:} \\ 
Key information
\end{tcolorbox}

\begin{tcolorbox}[
    title=P-CoT Reasoning subtask,
    boxrule=0.8pt,
    top=3mm,
    bottom=3mm,
    left=2mm,
    right=2mm,
    beforeafter skip=5mm]

\textbf{\small The input is:}\\
\small
\{System Prompt\} \\
Question \\
\{Information Retrieval Prompt\} \\ 
Key information \\ 
\{P-CoT Reasoning Prompt\} \\
\\
\textbf{\small The output is:} \\
P-CoT Reasoning Process
\end{tcolorbox}

The P-CoT reasoning solution is executed with a Python interpreter, and we get the {\textbf{P-CoT intermediate outputs}}. The format of P-CoT intermediate outputs is: \textbf{variable\_name1 = xxx, variable\_name2 = xxx}, etc. \textbf{Note} during the inference phase, for unexecutable P-CoT or variables without specific values, we use their variable names as intermediate results. In this time, the format of P-CoT intermediate outputs is: \textbf{variable\_name1 = xxx, variable\_name2, variable\_name3 = xxx}, etc.

\begin{tcolorbox}[
    title=Paradigm Conversion subtask,
    boxrule=0.8pt,
    top=3mm,
    bottom=3mm,
    left=2mm,
    right=2mm,
    beforeafter skip=5mm]

\textbf{\small The input is:}\\
\small
\{System Prompt\} \\
Question \\
\{Information Retrieval Prompt\} \\ 
{Key information} \\ 
\{P-CoT Reasoning Prompt\} \\
P-CoT Reasoning Process \\
The python solution’s intermediate outputs are: \{P-CoT intermediate outputs\} \\
\{Paradigm Conversion Prompt\}\\
\\
\textbf{\small The output is:} \\
N-CoT Reasoning Process
\end{tcolorbox}

\section{Training and implementation details}
\label{appendix:details}
\normalsize
We conduct all experiments with eight A100-80GB GPUs, and using DeepSpeed Zero stage 2 \cite{9355301, rasley2020deepspeed}, Huggingface accelerate \cite{accelerate} framework. We use AdamW \cite{DBLP:conf/iclr/LoshchilovH19} optimizer and set $eps$ to 1e-8.

\paragraph{Hyper parameters}
The maximum input length is set to 1024, while the maximum output length is 700. In SFT, the learning rate is 1e-5, the train batch size is 32, and no warm-up stage. We train models for 5 epochs, except for MathQA, for 10 epochs, and report the best performance. In RL, we use the best-initialized models in the SFT stage, the train batch size is 24, and we employ LORA in P-CoT RL and On-SL experiments for efficient and set $\alpha$ 64, $r$ 32. We set the policy and value learning rate 3e-7 for GSM8K and SVAMP and 1e-8 for MathQA. 
We set the partial reward $\epsilon$ to 0.1 and the convert penalty $\gamma$ to 0.2. The KL constraint coefficient $\beta$ is set to 0.05 for N-CoT experiments and 0.01 for P-CoT experiments. We train the model for 100 epochs and report the best performance. The discount factor and smoothing coefficient of GAE in PPO algorithm and the remaining details, we adhere to the settings in \cite{luong2024reft}.

\section{Dataset construction and sizes} \label{dataset}
The main details of datasets we used in this paper are presented in Table \ref{table:appendix_datasets}. The P-CoT annotations are derived from \cite{luong2024reft}. We use regular match (e.g., $re$ module in Python) for key information annotations based on P-CoT annotations according to the following principles:

\begin{itemize}
\item \small Due to the authenticity of P-CoT annotations, the variable names in the reasoning process signify their importance, which we use to identify relevant sentences in the question.
\item \small Sentences containing numbers and operators in the question.
\item \small If none of the above is available, we will take the questions that do not contain the question part as the key information.
\end{itemize}
After completing these steps, we performed manual validation.

For N-CoT annotations, we execute P-CoT and print the intermediate results as described in \ref{Subtask Subtasks}, then we use the GPT-4 \cite{DBLP:journals/corr/abs-2303-08774} for generation with the following prompts:

\begin{tcolorbox}[
    title=N-CoT Annotation Generation Prompt,
    boxrule=0.8pt,
    top=3mm,
    bottom=3mm,
    left=2mm,
    right=2mm,
    beforeafter skip=5mm]

\small
I will give you a math problem and a Python code: \\
\{P-CoT Reasoning Process\} \\
that solves this problem, along with the intermediate result information: \\
\{Intermediate Results\} \\
from this code. Please refer to the intermediate result information and python code to generate a natural language solution. The final answer should be given in the format \textbf{The answer is \textless answer\textgreater}. \\
\\
For example, if the final answer is 10, you should output \textbf{The answer is 10}. You don't need to output anything else; just output the natural language solution.
\end{tcolorbox}

We compared the final answer with the ground truth. For incorrect ones, after three attempts, those still unresolved were manually corrected. For all GPT-4-generated data, we used greedy decoding. Note due to SVAMP mostly consists of single-step solutions, we did not construct P-COT intermediate results for it.

\begin{table}[h!] 
  \centering
  \resizebox{1.0\linewidth}{!}{
    \begin{tabular}{llcc}
    \toprule
    \textbf{Category} & \textbf{Dataset} & \textbf{\#Train Samples} & \textbf{\#Test Samples} \\
    \midrule
    \multirow{3}{*}{Math Reasoning} & GSM8K \cite{Cobbe_Kosaraju_Bavarian_Hilton_Nakano_Hesse_Schulman_2021} & 7321 & 1319 \\
    & MathQA \cite{DBLP:conf/naacl/AminiGLKCH19} & 5706 & 1605 \\
    & SVAMP \cite{Patel_Bhattamishra_Goyal_2021} & 3052 & 1000 \\
    \midrule
  \end{tabular}
  }
  \caption{Statistics of the experiments datasets.}
  \label{table:appendix_datasets}
\end{table}

\newpage
\onecolumn

\section{Examples} \label{Example}
\subsection{Examples of Error Types}\label{example error}
In Section \ref{Pre}, we provide a detailed analysis of error types in N-CoT and P-CoT reasoning processes. For N-CoT, the predominant errors are comprehension errors (\textbf{CE}), calculation errors (\textbf{CA}), and logic inconsistencies (\textbf{LI}). For P-CoT, the main categories are reasoning errors (\textbf{RE}) and variable definition errors (\textbf{VE}).

\begin{figure*}[ht]
    \begin{tcbitemize}[raster columns=1, size=small]
        \tcbitem[title=N-CoT Reasoning Error Types, colback=white, colframe=blue!80!black]
        \begin{lstlisting}[style=plain, rulecolor=\color{red}, frame=single]
Comprehension Errors (CE)

Question: A robe takes 2 bolts of blue fiber and half that much white fiber. How many bolts in total does it take?
Ground true answer: 3

N-CoT SFT model output: 
Let's start by finding out how many bolts of white fiber the robe takes.
(*@\textcolor{red}{The white robe takes: bolts of white fiber = 2 bolts of blue fiber.}@*) 
Now, we can add the number of bolts of blue and white fiber together to find the total number of bolts the robe takes 2 bolts of blue fiber + 4 bolts of white fiber = 6 bolts in total. 
Therefore, the answer is 6.


        \end{lstlisting}
        \begin{lstlisting}[style=plain, rulecolor=\color{red}, frame=single]
Calculation Error (CA)

Question: A new program had 60 downloads in the first month. The number of downloads in the second month was three times as many as the downloads in the first month, but then reduced by 30% in the third month. How many downloads did the program have total over the three months?
Ground true answer: 366

N-CoT SFT model output: 
Let's break down the problem into three parts: First month: 60 downloads. Second month: 3 times as many as the downloads in the first month, which is 3 x 60 = 180 downloads.
Third month: 30% reduction in the number of downloads, 
(*@\textcolor{red}{which is 180 - (30/100 x 180) = 136 downloads.}@*)
To find the total number of downloads over the three months, we need to add up the number of downloads in each month: 60 + 180 + 136 = 376
Therefore, the answer is 376.


        \end{lstlisting}
        \begin{lstlisting}[style=plain, rulecolor=\color{red}, frame=single]
Logic Inconsistency (LI)

Question: a train 100 meters long completely crosses a 300 meters long bridge in 45 seconds. What is the speed of the train is?
Ground true answer: 32

N-CoT SFT model output: 
To solve this problem, we need to use the formula: 
Speed = Distance / Time 
First, we need to convert the time from seconds to meters:
(*@\textcolor{red}{45 seconds = 45 x 100 meters = 4500 meters}@*)
300 meters long bridge = 300 meters 
Now we can plug in the values: 
(*@\textcolor{red}{Speed = 4500 meters / 300 meters = 15 meters per second}@*)
Therefore, the answer is 15.


        \end{lstlisting}
    
    \end{tcbitemize}
\caption{The examples and analysis of Comprehension Error (\textbf{CE}), Calculation Error (\textbf{CA}) and Logic Inconsistency (\textbf{LI}) in N-CoT SFT model reasoning outputs.}
\label{fig:CE, CA and LI}
\end{figure*}

\newpage
\begin{figure*}[ht]
    \begin{tcbitemize}[raster columns=1, size=small]
        \tcbitem[title=P-CoT Reasoning Error Types, colback=white, colframe=green!60!black]
        \begin{lstlisting}[style=plain, rulecolor=\color{blue}, frame=single]
Reasoning Error (RE)

Question: a shopkeeper sold an article offering a discount of 5 % and earned a profit of 31.1 %. what would have been the percentage of profit earned if no discount had been offered?
Ground true answer: 38

P-CoT SFT model output: 
def solution():
    discount = 5    
    profit = 31.1   
    (*@\textcolor{red}{profit\_without\_discount = profit * (100 - discount) / 100}@*)
    result = profit_without_discount
    return result


        \end{lstlisting}
        \begin{lstlisting}[style=plain, rulecolor=\color{blue}, frame=single]
Variable Definition Error (VE)

Question: an art gallery has only paintings and sculptures. currently, 1 / 3 of the pieces of art are displayed, and 1 / 6 of the pieces on display are sculptures. if 1 / 3 of the pieces not on display are paintings, and 1000 sculptures are not on display, how many pieces of art does the gallery have?
Ground true answer: 2250

P-CoT SFT model output: 
def solution():
    (*@\textcolor{red}{pieces\_of\_art = 1 / 3}@*)
    (*@\textcolor{red}{piece\_on\_display = 1 / 6}@*)
    pieces_not_on_display = 1 - pieces_on_display
    sculptures_not_on_display = 1000
    pieces_total = pieces_of_art +  pieces_on_display + sculptures_not_on_display
    result = pieces_total
    return result


        \end{lstlisting}
    \end{tcbitemize}
\caption{The examples and analysis of Reasoning Error (\textbf{RE}) and Variable Definition Error (\textbf{VE}) in P-CoT SFT model reasoning outputs.}
\label{fig:RE and VE}
\end{figure*}

\newpage
\subsection{Examples Showcasing Gains from Parrot Training} \label{gains from parrot training}
As shown in Section \ref{Pre} and the Examples of Error Types \ref{example error}, N-CoT errors are primarily attributed to calculation errors and logical inconsistencies, while P-CoT errors largely arise from reasoning errors—issues that can be alleviated through Parrot hybrid training. In contrast, the intermediate results of P-CoT can mitigate the errors of N-CoT.

\begin{figure*}[ht]
    \begin{tcbitemize}[raster columns=1, raster equal height=rows, size=small]
        \tcbitem[title=Examples Showcasing Gains from Parrot Training,colback=white]
            \begin{lstlisting}[style=plain]
Question: Brian's friend Bobby has 5 fewer than 3 times as many video games as Brian does.  If Brian has 20 video games but lost 5 right before the comparison was made, how many does Bobby have?
Ground truth answer:40
        \end{lstlisting}
    \end{tcbitemize}
    \begin{tcbitemize}[raster columns=1, raster equal height=rows, size=small]
        \tcbitem[title=N-CoT SFT Model output, colback=white, colframe=blue!80!black]
        \begin{lstlisting}[style=plain, rulecolor=\color{red}, frame=single]
N-CoT SFT Model output:
We know that Bobby has 5 fewer than 3 times as many video games as Brian does. 
If Brian has 20 video games, then Bobby has:
(*@\textcolor{red}{3 x 20 = 60 video games, so Bobby has 60 video games}@*). 
Now we need to find out how many video games Brian had before he lost 5. 
(*@\textcolor{red}{If Brian had 20 video games and lost 5, then he had:20 - 5 = 15 video games.}@*) 
(*@\textcolor{red}{Therefore, Bobby has 60 - 5 = 55 video games}@*). 
The answer is: 55.
        \end{lstlisting}
        \tcbitem[title=Parrot SFT Model output,colback=white,colframe=green!60!black]
        \begin{lstlisting}[style=plain, rulecolor=\color{blue}, frame=single]
P-CoT Parrot SFT Model output:
def solution():   
    brian_games_initial = 20
    brian_games_lost = 5
    (*@\textcolor{green}{brian\_games\_final = brian\_games\_initial - brian\_games\_lost}@*)
    (*@\textcolor{green}{bobby\_games\_initial = brian\_games\_final * 3 - 5}@*)
    bobby_games_final = bobby_games_initial
    result = bobby_games_final
    return result
    
(*@\textbf{Parrot SFT P-CoT intermediate results: brian\_games\_final = 15, bobby\_games\_initial = 40.}@*)
        \end{lstlisting}
        \begin{lstlisting}[style=plain, rulecolor=\color{blue}, frame=single]
N-CoT Parrot SFT Model output:
Let's start by finding out how many video games Brian has after losing 5. 
(*@\textcolor{green}{Brian had 20 video games, but lost 5, so he now has 20 - 5 = 15 video games.}@*)
Next, we need to find out how many video games Bobby has. We know that Bobby has 5 fewer than 3 times as many video games as Brian, so we can set up an equation: 
(*@\textcolor{green}{Bobby's video games = 3 x Brian's video games - 5.}@*)
Substituting in the value we found for Brian's video games: 
Bobby's video games = 45-5=40
Therefore, Bobby has 40 video games. 
The answer is 40.
        \end{lstlisting}
    \end{tcbitemize}
\caption{The example and improvement analysis of the Parrot training pipeline. While the N-CoT SFT model exhibits logical errors, the guidance of P-CoT and its intermediate results ultimately led the model to correct the N-CoT solution.}
\label{fig:sft_rl_evolve}
\end{figure*}

\end{document}